\documentclass{article}
 
\usepackage{fullpage,cite}

\usepackage{graphicx} 
\graphicspath{{images/}}
\usepackage{subfigure}

\usepackage{titletoc}

\usepackage[utf8]{inputenc} 
\usepackage[T1]{fontenc}    
\usepackage{hyperref}       

\usepackage{url}            
\usepackage{booktabs}       
\usepackage{amsfonts}       
\usepackage{nicefrac}       
\usepackage{microtype}      
\usepackage{color}

\usepackage{multirow}
\usepackage{multicol}

\usepackage{algorithm}
\usepackage{algorithmic}

\usepackage{multibib}
\newcites{app}{Additional References}

\usepackage{bm,amsmath,amsthm,amssymb,color,array}
\usepackage{./packages/shortcuts}

\usepackage[textsize=footnotesize]{todonotes} 

\newcommand{\coint}[1]{\left[#1\right)}
\newcommand{\ocint}[1]{\left(#1\right]}

\title{On the Online Frank-Wolfe Algorithms for Convex and Non-convex Optimizations}

\author{Jean Lafond\thanks{Institut Mines-Telecom, Telecom ParisTech, CNRS LTCI, Paris, France. Email: \texttt{jean.lafond@telecom-paristech.fr}},~~~~Hoi-To Wai\thanks{School of Electrical, Computer and Energy Engineering, Arizona State University, AZ, USA. Email: \texttt{htwai@asu.edu}}~\thanks{J. Lafond and H.-T. Wai have contributed equally.},~~~~Eric Moulines\thanks{CMAP, Ecole Polytechnique, Palaiseau, France. Email: \texttt{eric.moulines@polytechnique.edu}}
}

\begin{document}

\maketitle

\begin{abstract}
In this paper, the online variants of the classical Frank-Wolfe algorithm are considered.
We consider minimizing the regret with a stochastic cost.
The  online algorithms only require
\emph{simple} iterative updates
and a \emph{non-adaptive} step size rule, in contrast to the \emph{hybrid} schemes
commonly considered in the literature.
Several new results are derived for convex and non-convex losses. 
With a strongly convex stochastic cost and when
the optimal solution lies in the interior of the constraint set \emph{or} the constraint set is a polytope,
the regret bound and anytime optimality 
are shown to be ${\cal O}( \log^3 T / T )$ and ${\cal O}( \log^2 T / T)$, 
respectively, where
$T$ is the number of rounds played.
These results are based on an improved analysis on the stochastic Frank-Wolfe algorithms. 
Moreover, the online algorithms are 
shown to converge even when the loss is \emph{non-convex}, i.e.,
the algorithms find a stationary point to the time-varying/stochastic loss at a rate of ${\cal O}(\sqrt{1/T})$.
Numerical experiments on realistic data sets are presented to support our theoretical claims.
\end{abstract}

\section{Introduction}
\label{sec:introduction}
Recently, Frank-Wolfe (FW) algorithm \cite{fw56} has become
popular for high-dimensional constrained optimization.
Compared to the projected gradient (PG) algorithm
(see \cite{Beck_Teboulle09,Juditsky_Nemirovski12a,Juditsky_Nemirovski12b,nemir09}),
the FW algorithm (a.k.a.~conditional gradient method) is appealing due to its \emph{projection-free} nature.
The costly projection step in PG is replaced by a linear optimization in FW.
The latter admits a closed
form solution for many problems of interests in machine learning.


This work focuses on the online variants of the FW and the
FW with \emph{away step} (AW) algorithms.
At each round, the proposed online FW/AW algorithms follow the same update equation applied in
classical FW/AW and
a step size is taken according to a non-adaptive rule.
The only modification involved is that we use an \emph{online-computed aggregated gradient} as a
surrogate of the true gradient
{of the} 
expected loss that we attempt to minimize.
We establish fast convergence  of the algorithms under various conditions.

Fast convergence for projection-free algorithms have been studied in \cite{simon_jaggi13,Lacoste_Jaggi15,garber_hazan_15,danhazan15,lan14,luo16}.
However, many works have considered a `hybrid' approach that involves solving a
regularized linear optimization during the updates \cite{danhazan15,lan14}; or combining existing
algorithms with FW \cite{luo16}.
In particular, the authors in \cite{danhazan15} showed a regret bound of
${\cal O}( \log T / T)$ for their online projection-free algorithm,
where $T$ is the number of {iterations}, under an adversarial setting. 
This matches the optimal bound for strongly
convex loss.
The drawback of these algorithms lies on the extra complexities (in implementation and
computation) added to the 
 classical FW algorithm.

Our aim is to show that simple online projection-free methods can achieve
on-the-par convergence guarantees as the sophisticated algorithms mentioned above.
In particular, we present a set of new  results
for online FW/AW algorithms under the full information setting, i.e.,
complete knowledge about the loss function is retrieved at each round \cite{linxiao10} (see \autoref{sec:problem-setup}).
Our online FW algorithm is similar to the online projection-free method proposed in
\cite{Hazan:Kale:12},
while the online AW algorithm is new.
For online FW algorithms,  \cite{Hazan:Kale:12} has proven
a regret of ${\cal O}( \sqrt{ \log^2 T / T } )$ for convex and smooth stochastic costs.
We improve the regret bound to ${\cal O} ( \log^3 T  / T)$ under
two different sets of assumptions: \textbf{(a)} the stochastic cost is strongly convex,
the optimal solutions lie in the interior of $\Cset$ (cf.~H\ref{iii3i}, for online FW);
\textbf{(b)} $\Cset$ is a polytope (cf.~H\ref{aw1}, for online AW).
An improved \emph{anytime} optimality bound of ${\cal O} (\log^2 T / T)$
(compared to ${\cal O}( \sqrt{ \log^2 T / T} )$ in \cite{Hazan:Kale:12}) is also proven.
We compare our results to the state-of-the-art in \autoref{tab:rate}.

\newcolumntype{C}[1]{>{\centering\let\newline\\\arraybackslash\hspace{0pt}}m{#1}}
\begin{table}[h]
\renewcommand{\arraystretch}{1.5}
\begin{center} {\small
\begin{tabular}{| c || C{3.9cm} | c | c |}
\hline
& Settings & Regret bound & \emph{Anytime} bound \\
\hline
\multirow{2}{*}{Garber \& Hazan, 2015 \cite{danhazan15}} & Hybrid algo., Lipschitz cvx.~loss & ${\cal O}( \sqrt{1/T} )$ & ${\cal O}( \sqrt{\log T/T})$ \tabularnewline
\cline{2-4}
& Hybrid algo., strong cvx.~loss & ${\cal O}( \log T / T )$ & ${\cal O}( {\log T/T})$ \tabularnewline
\hline
\multirow{2}{*}{Hazan \& Kale, 2012 \cite{Hazan:Kale:12}} & Simple algo., Lipschitz cvx.~loss & ${\cal O}(  \sqrt{\log^2 T/T} )$ & ${\cal O}( \sqrt{\log^2 T/T})$ \tabularnewline
\cline{2-4}
& Simple algo., strong cvx.~loss & ${\cal O}(  \sqrt{\log^2 T/T} )$ & ${\cal O}( \sqrt{\log^2 T/T})$ \tabularnewline
\hline
\multirow{2}{*}{This work} & Simple algo., strong cvx.~loss, interior~point (online FW)  & ${\cal O}( \log^3 T / T)$ & ${\cal O}( \log^2 T / T )$ \tabularnewline
\cline{2-4}
& Simple algo., strong cvx.~loss, polytope const.~(online AW)  & ${\cal O}( \log^3 T / T)$ & ${\cal O}( \log^2 T / T )$ \tabularnewline
\hline
\end{tabular}} \vspace{.2cm}
\caption{Convergence rate comparison.
Note that the regret bound for \cite{danhazan15} is given under an adversarial loss setting,
while the bounds
for \cite{Hazan:Kale:12} and our work are based on a stochastic cost.
Depending on the applications
(see Section~\ref{sec:num} \& \autoref{sec:olasso}),
our regret and anytime bounds can be improved to ${\cal O}( \log^2 T / T)$ and ${\cal O} (\log T / T)$,
respectively.} \label{tab:rate} \vspace{-.4cm}
\end{center}
\end{table}

Another interesting discovery is that the online FW/AW algorithms converge to a stationary point
even when the loss is \emph{non-convex}, at a rate of ${\cal O} (1/\sqrt{T})$. To the best of our 
knowledge, this is the first convergence rate
result for non-convex online optimization with projection-free methods.

To support our claims, we perform numerical experiments on online matrix completion using realistic
dataset. The proposed online schemes outperform
a simple projected gradient method in terms of running time. The algorithm also demonstrates excellent performance
for robust binary classification.

\textbf{Related Works}. 
In addition
 to the references mentioned above,
this work is related to the study of stochastic optimization, e.g., \cite{henry15,nemir09}.
\cite{henry15} describes a FW algorithm using stochastic approximation and proves that the optimality
gap converges to zero almost surely;
\cite{nemir09} analyses the stochastic projected gradient method and proves that the convergence rate
is ${\cal O}( \log t / t )$ under strong convexity and that the optimal solution
lies in the interior of $\Cset$. This is similar to assumption H\ref{iii3i} in this paper.

Lastly, most recent works on \emph{non-convex} optimization are based on the
stochastic projected gradient descent method \cite{zhu16,rong15}.
Projection-free \emph{non-convex} optimization has only been addressed by
a few authors \cite{henry15,serbia}.
At the time when we finished with the writing, 
we notice that several authors have published articles pertaining to offline, non-convex FW algorithm,
e.g., \cite{simon16} achieves the same convergence rate as ours with an adaptive step size,
\cite{jiang16} considers a different assumption on the smoothness of loss function,
\cite{yu14} has a slower convergence rate than ours. 
Nevertheless, none of the above has considered an online optimization setting 
with time varying objective like ours.

\textbf{Notation}. For any $n \in \NN$,
 let $[n]$ denote
 the set $\{1,\cdots,n\}$.
 The inner product
on a $n$ dimensional real Euclidian space $\E$ is denoted by $\langle \cdot,\cdot \rangle$
and the associated Euclidian norm by $\| \cdot \|_2$.
The space $\E$ is also equipped with a norm
$\|\cdot\|$ and its dual norm $\|\cdot\|_{\star}$.
Diameter of the set $\Cset$ \wrt $\|\cdot\|_\star$
is denoted by $\rho$, that is
$\textstyle \rho \eqdef \sup_{\theta,\theta' \in \Cset} \| \theta - \theta' \|_\star$.
In addition, we denote the diameter of $\Cset$ \wrt the Euclidean norm as $\bar{\rho}$, i.e., $\bar{\rho} \eqdef
\sup_{\theta,\theta' \in \Cset} \| \theta - \theta' \|_2$.
The $i$th element in a vector ${\bm x}$ is denoted by $[{\bm x}]_i$.

\section{Problem Setup and Algorithms}
\label{sec:problem-setup}
We use the setting introduced in \cite{Hazan:Kale:12}.
The online learner wants to minimize a loss function $f$
which is the expectation of empirical loss functions $f_t (\prm) = f (\prm; \omega_t )$,
where $\omega_t$
is drawn i.i.d.~from a fixed distribution ${\cal D}$: $\Obj (\prm) \eqdef \EE_{\omega \sim {\cal D}} [ \Obj ( \prm ; \omega ) ]$.
The regret of a sequence of actions $\{\prm_t\}_{t=1}^T$ is~:
\beq \label{eq:regret} \textstyle
{\cal R}_T \eqdef T^{-1}  \sum_{t=1}^T f(\prm_t)  - \min_{\prm \in \Cset} f(\prm) \eqs.
\eeq
Here,  $\Cset$ is a bounded convex set included in
$\E$ and $f_t(\cdot)$ is a continuously differentiable function.

Our proposed algorithms assume the \emph{full information} setting \cite{linxiao10}
such that upon playing $\prm_t$,
we receive full knowledge about the loss function $\prm \mapsto f_t( \prm)$ . The choice of $\prm_{t+1}$
will be based on the previously observed loss $\{ f_s (\prm ) \}_{s=1}^t$. Let $\gamma_t \in (0,1]$ be a
sequence of decreasing step size (see \autoref{sec:main}),
$F_T(\prm) = \sum_{t=1}^T f_t(\prm)$ the aggregated loss and
$\grd F_T(\prm)$ be the gradient
of $F_t(\prm)$ evaluated at $\prm$,
we study two online algorithms.

\begin{multicols}{2}{
\textbf{Online Frank-Wolfe (O-FW).}
The online FW algorithm, introduced in \cite{Hazan:Kale:12},  is a direct generalization of the classical FW algorithm, as summarized in \autoref{alg:ofw}. It differs from the classical FW algorithm only in the sense that
the \emph{aggregated} gradient $\grd F_t(\prm_t) = t^{-1} \sum_{s=1}^t \grd f_s (\prm_t)$
is used for the linear optimization in Step~\ref{ofw:lo}.
See the comment in Remark~\ref{rem@complexity-iteration} for the complexity of
calculating the aggregated gradient.

\algsetup{indent=1em}
\begin{algorithm}[H]
\caption{Online Frank-Wolfe (O-FW).}\label{alg:ofw}
  \begin{algorithmic}[1]
  \STATE \textbf{Initialize:} ${{\prm}}_1 \leftarrow 0$
  \FOR {$t=1,\dots$}
  \STATE Play $\prm_t$ and receive $\prm \mapsto f_t (\prm)$.
   \STATE Solve the linear optimization: \label{ofw:lo}
   \begin{equation} \label{eq:linear}
   {\atom}_t \leftarrow \arg \min_{ {\atom} \in \Cset} ~\big\langle {\atom}, \grd F_t (\prm_t) \big\rangle \eqs. \vspace{-.2cm}
   \end{equation}
   \STATE Compute ${\prm}_{t+1} \leftarrow {\prm}_t + \gamma_t ({\atom}_t - {\prm}_t)$.
\ENDFOR
  \end{algorithmic}
\end{algorithm}}
\end{multicols}

\algsetup{indent=1em}
\begin{algorithm}[t]
\caption{Online away step Frank-Wolfe (O-AW).}\label{alg:ofw_away}
  \begin{algorithmic}[1]
  \STATE \textbf{Initialize:} $\rstep{0} = 0$, ${{\prm}}_1 = 0$,  $\activ_1 = \emptyset$;
  \FOR {$t=1, \dots$}
  \STATE Play $\prm_t$ and receive the loss function $\prm \mapsto f_t(\prm)$ .
   \STATE Solve the linear optimizations with the aggregated gradient:
   \beq \label{eq:away}
   \atom_t^{\sf FW} \leftarrow
   \arg \min_{ {\atom} \in \Cset} ~\big\langle {\atom}, \grd F_t ({\prm}_t) \big\rangle,~ \atom_t^{\sf AW} \leftarrow \arg \max_{ {\atom} \in \activ_t } ~\big\langle {\atom}, \grd F_t ({\prm}_t) \big\rangle \vspace{-.4cm}
   \eeq
   \IF {$\langle \atom_t^{\sf FW} - \prm_t, \grd F_t ({\prm}_t)
   \rangle \leq \langle \prm_t-\atom_t^{\sf AW} , \grd F_t ({\prm}_t) \rangle$
   or $\activ_t = \emptyset$} \label{ln:cond}
   \STATE \texttt{FW step:}
   ${\bm d}_t \leftarrow \atom_t^{\sf FW} - \prm_t$,
   $\rstep{t} \leftarrow \rstep{t-1}+1$, $\hat{\gamma}_t \leftarrow \gamma_{\rstep{t}}$ and
   $\activ_{t+1} \leftarrow \activ_{t} \cup \{\atom_t^{\sf FW}\}$.
   \label{ln:fw}
   \ELSE
   \STATE
   ${\bm d}_t \leftarrow \prm_t -\atom_t^{\sf AW}$,
   $\gamma_{\rm max} =  \alpha_t^{\atom_t^{\sf AW}} / (1 - \alpha_t^{\atom_t^{\sf AW}})$,
   cf.~\eqref{eq:def_active} for definition of $\alpha_t^{\atom_t^{\sf AW}}$.\label{ln:away}
   \IF{ $\gamma_{\rm max} \geq  \gamma_{\rstep{t-1}}$} \label{ln:drop_cond}
   \STATE  \texttt{AW step:} $\rstep{t} \leftarrow \rstep{t-1}+1$ and $\hat{\gamma}_t \leftarrow  \gamma_{\rstep{t}}$\label{ln:aw}
   \ELSE
   \STATE \texttt{Drop step:} $\hat{\gamma}_t \leftarrow \gamma_{\rm max}$, $\rstep{t} \leftarrow \rstep{t-1}$ and
   $\activ_{t+1} \leftarrow \activ_{t} \setminus \{\atom_t^{\sf AW}\}$  \label{ln:drop}
   \ENDIF
   \ENDIF
   \STATE Compute ${\prm}_{t+1} \leftarrow {\prm}_t + \hat{\gamma}_t {\bm d}_t$.\label{ln:move}
\ENDFOR
  \end{algorithmic}
\end{algorithm}

\textbf{Online away-step Frank-Wolfe (O-AW)}.
The online counterpart of the away step algorithm is given in \autoref{alg:ofw_away}.
By construction, the iterate $\prm_t$ is
a convex combination of extreme points of $\Cset$, referred to as active atoms.
We denote by $\activ_t$ the set of active atoms and denote by $\alpha_t^\atom$
the positive weight of any active atom $\atom \in \activ_t$ at time $t$, that is:
\begin{equation}
 \label{eq:def_active}
\textstyle \prm_t = \sum_{ \atom \in \activ_t} \alpha_t^{\atom} \cdot \atom \quad \text{with} \quad \alpha_t^\atom >0 \eqs.
\end{equation}

At each round, two types of step might be taken.
If the condition of line \ref{ln:cond} in \autoref{alg:ofw_away} is satisfied,
we call the iteration a ``FW step'', otherwise
we call it an ``AW step''.
When a FW step is taken,
a new atom $\atom_t^{\sf FW}$ is selected \eqref{eq:away}, the current iterate
$\prm_t$ is moved towards $\atom_t^{\sf FW}$
and the active set
is updated accordingly (lines \ref{ln:fw} and \ref{ln:move}). The selected
atom is the (extreme) point of $\Cset$ which is maximally
correlated to the negative aggregated gradient.
Note that this step is identical to a usual O-FW iteration.
When an ``AW step'' is taken, a currently active atom
$\atom_t^{\sf AW}$ is selected \eqref{eq:away}
and the current iterate is moved away from
$\atom_t^{\sf AW}$ (line \ref{ln:away} and \ref{ln:move}).
The atom $\atom_t^{\sf AW}$ is the active atom
which is the most correlated to the current gradient approximation.
The intuition is that taking the `away' step prevents the algorithm from following a `zig-zag' path when $\prm_t$
is close to the boundary of $\Cset$ \cite{Wolfe70}.

Lastly, we note that the O-AW algorithm is similar to a classical AW algorithm \cite{Wolfe70}.
The exception is that a fixed step size rule is adopted due
to the online optimization setting.

\begin{Remark} \label{rem:atom}
As the linear optimization \eqref{eq:away} enumerates over the active atoms
${\cal A}_t$ at round $t$, the O-AW algorithm is suitable when
$\Cset$  is an atomic (or polytope) set,
otherwise $|{\cal A}_t|$ may become too large.
\end{Remark}

\begin{Remark}[Linear Optimization.]
The run-time complexity of the O-FW  and O-AW algorithms depends
on finding efficient solution to the linear optimization step.
In many cases, this is extremely efficient.
For example,  when $\Cset$ is the trace-norm ball, then the linear optimization
amounts to finding the top singular vectors of the gradient; see~\cite{Jaggi13} for an overview.
\end{Remark}

\begin{Remark}[Complexity per iteration.]
\label{rem@complexity-iteration}
In addition to the linear optimization,
both O-FW/O-AW algorithms require the aggregate gradient $\grd F_t(\prm_t)$ to be computed
at each round, and the complexity involved grows with the round number.
In cases when the loss $f_t$ is the negated log-likelihood of an exponential
family distribution, the gradient aggregation can be replaced by an efficient `on-the-fly' update,
whose complexity is a dimension-dependent \textbf{constant over the iterations}.
As demonstrated in Section~\ref{sec:num} and \autoref{sec:olasso}, this set-up covers  many problems of interest, among others the online matrix completion and online LASSO.
\end{Remark}


\section{Main Results}
\label{sec:main}
This section presents the main results for the convergence of O-FW/O-AW algorithms.
Notice that our results for convex losses are based on an improved analysis 
on the \emph{stochastic/inexact} invariant of FW/AW algorithms 
(see \emph{Anytime Analysis} in \autoref{sec:cvx}), while 
the results for non-convex losses are derived from a novel observation 
on the duality gap for FW algorithms. 
Due to space constraints, only the main results are displayed.
Detailed proofs can be found in the appendices.

Some constants are defined as follows. 
A function $f$ is said to be $\mu$-strongly convex if, for all $\prm$, $\tilde{\prm} \in \E$,
\begin{equation}
\label{eq:mu}
\Obj (\prm) \leq \Obj (\tilde{\prm}) +
\langle \grd \Obj (\prm), \prm - \tilde{\prm} \rangle
- (\mu / 2) \| \prm - \tilde{\prm} \|_{{2}}^2\eqsp.
\end{equation}
We also say $f$ is $L$-smooth if for all $\prm$, $\tilde{\prm} \in \E$ we get
\begin{equation}
\label{eq:L}
\Obj (\tilde{\prm}) \leq \Obj (\prm) +
\langle \grd \Obj (\prm), \tilde{\prm} - \prm \rangle
+ (L/2) \| \prm - \tilde{\prm} \|_{{2}}^2  \eqsp.
\end{equation}
Lastly, $f$ is said to be $G$-Lipschitz if for
all $\prm$, $\tilde{\prm} \in \E$,
\beq
| f( \prm) - f (\tilde{\prm} ) | \leq G \| \prm - \tilde{\prm} \|_* \eqsp.
\eeq

\subsection{Convex Loss} \label{sec:cvx}
We analyze first \autoref{alg:ofw} and \autoref{alg:ofw_away}
when the expected loss function $f$ is convex.
In particular, our analysis will depend on the following geometric condition of the constraint set $\Cset$.
Denote by $\partial \Cset$ the boundary set of $\Cset$.
For \autoref{alg:ofw}, we consider
\begin{assumption}\label{iii3i}
There is a minimizer $\prm^\star$ of $f$ that lies in the interior of $\Cset$, i.e.,
$\delta \eqdef \inf_{ { \bm s} \in \partial \Cset} \| {\bm s} - \prm^\star \|_2 > 0$.
\end{assumption}
While H\ref{iii3i} appears to be restrictive, for \autoref{alg:ofw_away}, we can work with a relaxed
condition:
\begin{assumption}\label{aw1}
$\Cset$ is a polytope. 
\end{assumption}
As argued in \cite{Lacoste_Jaggi15}, 
H\autoref{aw1} implies that the pyramidal width for $\Cset$, $\daw \eqdef PdirW( \Cset )$, is positive; 
see the definition in \eqref{eq:pywidth} of the appendix. \vspace{.2cm}

\noindent \textbf{Regret Analysis.}
Our main result is summarized as follows. For $\epsilon \in (0,1)$,

\begin{Theorem} \label{cor:anytime}
Consider O-FW (\resp O-AW). Assume H\ref{iii3i} (\resp H\ref{aw1}),
$f(\prm)$ is $\mu$-strongly convex, 
$f(\prm; \omega)$ is $L$-smooth for all $\omega$ drawn from ${\cal D}$
and each element of $\grd f_t (\prm)$ is sub-Gaussian with parameter $\sigma_D$.
Set $\gamma_t = 2 / (t+1)$.
With probability at least $1-\epsilon$ and for all $t \geq 1$, the anytime loss bounds hold:
\beq \label{eq:anytime_bd}
\begin{split}
\text{(O-FW)} &~~~~f(\prm_t) - \min_{\prm \in \Cset} f(\prm) \leq \big( 2 \sqrt{3/2} (\sigma_{grd} \rho + \Curve) / (2 \delta \sqrt{\mu}) \big)^2 \cdot ( \log(t) \log(nt/\epsilon)  )\cdot t^{-1} \eqsp,\\
\text{(O-AW)} &~~~~f(\prm_t) - \min_{\prm \in \Cset} f(\prm) \leq \big( (5/3) (2 \sigma_{grd} \rho + \Curve) / (\daw \sqrt{\mu}) \big)^2 \cdot ( \log(t) \log(nt/\epsilon)  )\cdot t^{-1}  \eqsp,
\end{split}
\eeq
where $\sigma_{grd} = {\cal O} ( \max\{ \sigma_D, \bar\rho L \} \sqrt{n} )$. 
Consequently, summing up the two sides of \eqref{eq:anytime_bd}
from $t=1$ to $t=T$ gives the regret bound for both O-FW and O-AW:
\beq \textstyle
T^{-1} \sum_{t=1}^T f(\prm_t) - \min_{\prm \in \Cset} f(\prm) = {\cal O} ( \log^3 T / T ),~\forall~T \geq 1 \eqs.
\eeq
\end{Theorem}

\begin{proof}
To prove Theorem~\ref{cor:anytime},
we first upper bound the gradient error of $\grd F_t (\prm_t)$, i.e.,
\begin{Prop} \label{prop:ogrd}
Assume that $f(\prm; \omega)$ is $L$-smooth for all $\omega$ from ${\cal D}$ and each element 
of the vector $\grd f_t (\prm)$ is sub-Gaussian with 
parameter $\sigma_D$. With probability at least $1- \epsilon$,\vspace{-.1cm}
\beq
\| \grd F_t( \prm_t ) - \grd f( \prm_t ) \|_\infty = {\cal O}( \max\{ {\sigma_D} , \bar\rho L \}  \sqrt{n}\sqrt{ {\log(t) \log(n t/\epsilon) }/{t}} ),~\forall~t \geq 1 \eqs.
\eeq
\end{Prop}
This shows that $\grd F_t(\prm_t)$ is an \emph{inexact} gradient of the stochastic objective $f(\prm)$
at $\prm_t$.
Our proof is achieved by applying Theorem \autoref{thm:sfw} (see below) by plugging in the appropriate constants.
\end{proof} \vspace{-.2cm}

We notice that for O-FW, \cite{Hazan:Kale:12} has proven a regret bound of 
${\cal O}( \sqrt{ \log^2 T / T } )$, which is obtained by applying a  
uniform approximation bound on the \emph{objective value} and proving a ${\cal O}(1/\sqrt{t})$
bound for the instantaneous loss $F_t( \prm_t) - F_t(\prm_t^\star)$.
In contrast, Theorem~\ref{cor:anytime} yields an improved regret by controlling the
\emph{gradient error} directly using 
Proposition~\ref{prop:ogrd} and analyzing O-FW/O-AW as an FW/AW algorithm with \emph{inexact} 
gradient in the following. \vspace{.2cm}

\noindent \textbf{Anytime Analysis.} 
The regret analysis is derived from the following
general result for FW/AW algorithms with \emph{stochastic/inexact} gradients.
Let $\Hgrd_t f(\prm_t)$ be an estimate of $\nabla f(\prm_t)$
which satisfies:
\begin{assumption}\label{iii1}
For some $\alpha \in \ocint{0,1}$, $\sigma \geq 0$ and $K \in \mathbb{Z}_+^\star$. With probability at least $1- \epsilon$, we have
\beq
\| \Hgrd_t f( \prm_t) - \grd f(\prm_t) \| \leq \sigma ( \noise_t / \{ K+t-1 \} )^\alpha,~\forall~t \geq 1\eqs,
\eeq
where $\noise_t \geq 1$ is an increasing sequence such that the right hand side decreases to 0.
\end{assumption}
This is a more general setting than is required for the analysis of O-FW/O-AW 
as $\sigma,\alpha, \noise_t$ are arbitrary. 
The O-FW (or O-AW) with the above inexact gradient has the following convergence rate:
\begin{Theorem} \label{thm:sfw}
Consider the sequence $\{ \prm_t \}_{t=1}^\infty$ generated by O-FW
(\resp O-AW) with the aggregated gradient
$\grd F_t (\prm_t)$ replaced by $\Hgrd_t f(\prm_t)$ satisfying H\ref{iii1} with $K=2$.
Assume H\ref{iii3i} (\resp H\ref{aw1}) and that
$f(\prm)$ is $L$-smooth, $\mu$-strongly convex. Set $\gamma_t = 2 / (t+1)$. 
With probability at least $1-\epsilon$ and for all $t \geq 1$, we have \vspace{-.1cm}
\beq
\begin{split}
\text{(O-FW)} &~~~f(\prm_t) - \min_{\prm \in \Cset} f(\prm) \leq \big(\max\{ 2 (3/2)^{\alpha}, 1 + 2\alpha / (2-\alpha) \} (\sigma \rho + \Curve) / (2 \delta \sqrt{\mu}) \big)^2 \cdot (\noise_t / (t+1))^{2 \alpha} \eqsp, \\
\text{(O-AW)} &~~~f(\prm_t) - \min_{\prm \in \Cset} f(\prm) \leq 2 \big( \max\{ (3/2)^\alpha, 1 + 2\alpha / (2-\alpha) \} ( 2 \sigma \rho + \Curve ) / ( \daw \sqrt{\mu} ) \big)^2 \cdot (\noise_t / (t+1))^{2 \alpha} \eqsp,
\end{split}
\eeq
\end{Theorem}

When $\alpha = 0.5$, Theorem~\ref{thm:sfw} improves the previous known bound of
$f(\prm_t) - \min_{\prm \in \Cset} f(\prm) = {\cal O} ( \sqrt{ \noise_t / t } )$
in \cite{freund_grigas,Jaggi13} under strong convexity and H\ref{iii3i} or H\ref{aw1}.
It also matches the information-theoretical lower bound for strongly convex stochastic
optimization in \cite{maxim11} (up to a log factor).
Moreover, for O-AW, the strong convexity requirement on $f$
can be relaxed; see Appendix~\ref{sec:pf_nonstrong}.

\subsection{Non-convex Loss}
Define respectively the \emph{duality gaps} for O-FW and O-AW as
\beq \label{eq:dual}
g_t^{\sf FW} \eqdef \langle \grd F_t(\prm_t),  \prm_t - \atom_t \rangle,~~
 g_t^{\sf AW} \eqdef \langle \grd F_t(\prm_t), \atom_t^{\sf AW} - \atom_t^{\sf FW} \rangle\eqs,
\eeq
where $\atom_t$ is defined in line~\ref{ofw:lo} of \autoref{alg:ofw} and $\atom_t^{\sf AW}, \atom_t^{\sf FW}$ are defined in \eqref{eq:away} of \autoref{alg:ofw_away}.
Using the definition of $\atom_t$, if $g_t^{\sf FW} = 0$, 
then $\prm_t$ is a stationary point to the optimization
problem $\min_{ \prm \in \Cset } F_t (\prm)$. 
Therefore, $g_t^{\sf FW}$ (and similarly $g_t^{\sf AW}$) 
can be seen as a measure to the stationarity of the point $\prm_t$
to the online optimization problem. 

We analyze the convergence of O-FW/O-AW for general Lipschitz and smooth (possibly non-convex) loss function using the duality gaps defined above.
To do so, we depart from the usual induction based proof technique (e.g., in the previous section
or \cite{Jaggi13,Hazan:Kale:12}).
Instead, our method of proof amounts  to relate the duality gaps with a learning rate controlled
by the step size rule on $\gamma_t$.
The main result can be found below:

\begin{Theorem} \label{thm:dual}
Consider O-FW and O-AW. Assume that each of the loss function $f_t$ is $G$-Lipschitz, $L$-smooth.
Setting the step size sequence as $\gamma_t = t^{-\alpha}$ with $\alpha \in \coint{0.5,1}$. We have \vspace{-.1cm}
\beq \label{eq:dual_bd}
\begin{split}
\min_{ t \in [T/2+1,T] } g_t^{\sf FW} & \leq (1-\alpha) ( 4 G \rho + \Curve/2) (1-(2/3)^{1-\alpha})^{-1} \cdot T^{-(1-\alpha)},~\forall~T \geq 6  \eqs, \\
\min_{ t \in [T/2+1,T] } g_t^{\sf AW} & \leq (1-\alpha) ( 4 G \rho + \Curve) (1-(4/5)^{1-\alpha})^{-1} \cdot T^{-(1-\alpha)},~\forall~T \geq 20 \eqs.
\end{split}
\eeq
\end{Theorem}

Notice that the above result is deterministic (cf.~the definition of $g_t^{\sf FW}$, $g_t^{\sf AW}$)
and also works with non-stochastic, non-convex losses.
The above guarantees an ${\cal O}( 1 / T^{1-\alpha} )$ rate for
O-FW/O-AW at a certain round $t$ within the interval $[T/2+1, T]$.
Unlike the regret/anytime analysis done previously,
our bounds are stated with respect to
the \emph{best} duality gap attained within an interval from $t=T/2+1$ to $t=T$.
This is a common artifact when analyzing the duality
gap of FW \cite{Jaggi13}.
Furthermore, we can show that:

\begin{Prop} \label{cor:ncvx}
Consider O-FW (or O-AW), assume
that each of $f_t$ is $G$-Lipschitz, $L$-smooth 
and each of $\grd f_t (\prm)$ is sub-Gaussian with parameter $\sigma_D$.
Set the step size sequence as $\gamma_t = t^{-\alpha}$ with $\alpha \in [0.5,1)$.
With probability at least $1-\epsilon$ and for $T \geq 20$, there exists $t \in [T/2+1,T]$ such that
\beq
\max_{ \prm \in \Cset }~\langle \grd f(\prm_t), \prm_t - \prm \rangle = {\cal O}\big( \max \big\{ 1 / T^{1-\alpha}, \sqrt{ \log T / T } \big\} \big) \eqs.
\eeq
\end{Prop}
The proposition indicates that the iterate $\prm_t$ at round $t \in [T/2+1,T]$ is an
${\cal O}\big( \max \big\{ 1 / T^{1-\alpha}, \sqrt{ \log T / T } \big\} \big)$-stationary
point to the stochastic optimization $\min_{\prm \in \Cset} f (\prm)$.
Our proof relies on Theorem~\ref{thm:dual} and a uniform approximation bound
result for $\grd F_t( \prm_t)$. 

%

\section{Sketch of the Proof of Theorem~\ref{thm:sfw}}
To provide some insights, we present
the main ideas behind the proof of Theorem~\ref{thm:sfw}.
To simplify the discussion we only consider O-FW, $K=1$, $\noise_t = 1$ and $\alpha = 0.5$
in  H\ref{iii1}. The full proof can be found in the supplementary material.
Since $\Obj(\cdot)$ is $L$-smooth and $\Cset$ has a diameter of $\bar{\rho}$,
we have
\begin{equation*}
 \Obj ( \prm_{t+1} )  \leq \Obj ( \prm_t ) + \gamma_t \langle \grd \Obj (\prm_t), \atom_t - \prm_t \rangle +  \gamma_t^2 \Curve / 2
\end{equation*}
If we define $\bm{\epsilon}_t \eqdef \Hgrd_t f( \prm_t) - \grd f(\prm_t)$, and subtract $\Obj ( \prm^* )$ on both sides,
applying Cauchy Schwartz yields
\begin{equation}
\label{eq:sketch}
h_{t+1} \leq h_{t} - \gamma_t g_t^{\sf FW} +  \gamma_t^2 \Curve/2 +  \gamma_t {\rho} \| \bm{\epsilon}_t \| \eqs.
\end{equation}
Observe that as $h_t, g_t^{\sf FW} \geq 0$, the duality gap term
$g_t^{\sf FW}$ determines the convergence rate of the sequence $h_t$ to zero.

In fact, when $\Obj$ is convex, one can prove $g_t^{\sf FW}\geq h_t - {\rho} \| \bm{\epsilon}_t \|$. 
By the assumption  H\ref{iii1}, with probability at least $1-\epsilon$, we have
\begin{equation*}
h_{t+1} \leq h_{t} - \gamma_t h_t + \gamma_t^2 \Curve / 2 + 2 \gamma_t {\rho} {\sigma} / {\sqrt{t}}
= (1 - \gamma_t) h_t + {\cal O} (t^{-1.5}) \eqs.
\end{equation*}
Setting $\gamma_t = 1/t$ and a simple induction on the above inequality 
proves $h_t = \mathcal{O}(1/\sqrt{t})$.

An important consequence of H\ref{iii3i} is that the latter leads to a tighter lower
bound on $g_t^{\sf FW}$. 
As we present in Lemma~\ref{lem:stg_cvx} in \autoref{sec:regretpf}, under H\ref{iii3i} and when $f$ is 
$\mu$-strongly convex, we can lower bound $g_t^{\sf FW}$ as  
\[
g_t^{\sf FW}\geq \max\{0, \delta \sqrt{\mu h_t} - {\rho} \| \bm{\epsilon}_t \|\}.
\]
Note that $h_t$ converges to zero and the above lower bound on $g_t^{\sf FW}$
eventually will become tighter than the previous one, 
i.e., $g_t^{\sf FW} \geq \delta \sqrt{\mu h_t} - {\rho} \| \bm{\epsilon}_t \|
\geq h_t - {\rho} \| \bm{\epsilon}_t \|$. 
This leads to the accelerated convergence of $h_t$. More formally,
 plugging the lower bound into \eqref{eq:sketch} gives
 \begin{equation*}
h_{t+1} \leq h_{t} - \gamma_t \delta \sqrt{\mu h_t} + \gamma_t^2 \Curve / 2 + 2 \gamma_t {\rho} {\sigma}/ {\sqrt{t}}\eqs.
\end{equation*}
Again, setting $\gamma_t = 1/t$ and a carefully executed induction 
argument shows $h_t = \mathcal{O}(1/t)$. The same line of arguments
 is also used to prove the convergence rate of O-AW, where H\ref{aw1} will be 
 required (instead of H\ref{iii3i}) to provide 
 a similarly tight lower bound to $g_t^{\sf AW}$. 

\section{Numerical Experiments}
\label{sec:num}
We conduct numerical experiments to demonstrate the practical performance of the online algorithms. 
An additional experiment for online LASSO with O-AW can be found in the appendix.

\subsection{Example: Online matrix completion (MC)} 
Consider the following setting: 
we are sequentially  
given observations   
in the form $(k_{t},l_{t},\Y_{t})$, with $(k_{t}, l_{t}) \in [m_1]\times [m_2]$ 
and $\Y_{t} \in \RR$. 
The observations are assumed to be i.i.d.
To define the loss function, the conditional
distribution  of $\Y_{t}$ \wrt the sampling
is parametrized by
an unknown matrix $\tX \in \matset{m_1}{m_2}$
and supposed to belong to the exponential family, i.e.,
\beq 
p_{\tX}(\Y_t | k_t, l_t ) \eqdef
 \bms(\Y_{t}) \exp\left(\Y_{t} \tX_{k_{t},l_{t}} - \lgp(\tX_{k_{t},l_{t}}) \right),
 \eeq
where $\bms(\cdot)$ and $\lgp(\cdot)$ are the base measure and log-partition functions,
respectively.
A natural choice for the loss function at round $t$ 
is obtained by taking the logarithm of the posterior, i.e.,
\[
f_t (\prm) \eqdef \lgp(\prm_{k_{t},l_{t}})- \Y_{t} \prm_{k_{t},l_{t}}.
\]
Our goal is to minimize 
the regret 
with a penalty favoring low rank solutions $\Cset \eqdef \{ \prm \in \matset{m_1}{m_2} : 
\| \prm \|_{\sigma,1} \leq \trad \}$,
and the stochastic cost associated is $\Obj(\prm) \eqdef  \EE_{\bar{\prm}} 
\big[  \lgp(\prm_{k_{1},l_{1}})- \Y_{1} \prm_{k_{1},l_{1}} \big]$.

Note that the aggregated gradient $\grd F_t (\prm_t) = t^{-1} \grd \sum_{s=1}^t f_s(\prm_t)$
can be expressed as:
\beq \notag \textstyle
\big[ \grd F_t (\prm_t) \big]_{k,l} =   t^{-1}  \lgp'( [\prm_t]_{k,l} ) \big[ \sum_{s=1}^t e_{k_s} e_{l_s}^{'\top} \big]_{k,l} - t^{-1} \big[ \sum_{s=1}^t Y_s e_{k_s} e_{l_s}^{'\top} \big]_{k,l},~\forall~k,l \in [m_1] \times [m_2] \eqs,
\eeq
with $\{e_k\}_{k=1}^{m_1}$ (\resp $\{e'_l\}_{l=1}^{m_2}$) the canonical basis of
of $\RR^{m_1}$ (\resp $\RR^{m_2}$).
We observe that the two matrices $\sum_{s=1}^t e_{k_s} e_{l_s}^{'\top}$ and 
$\sum_{s=1}^t Y_s e_{k_s} e_{l_s}^{'\top}$ can be computed `on-the-fly' as the running sum. 
The two matrices can also be stored efficiently in the memory as they are at most $t$-sparse. 
The per iteration complexity is upper bounded by ${\cal O}(\min\{m_1m_2,T\})$, where $T$ is the total number of observations. 

We observe that for online MC, a better anytime/regret bound than the general case
analyzed in Section~\ref{sec:main} can be achieved. In particular, 
\autoref{prop:mc_simp} shows that
$
\| \grd F_t (\prm) - \grd f(\prm) \|_{\sigma, \infty} = {\cal O} ( \sqrt{ \log t / t } )
$.
As such, the online gradient satisfies H\ref{iii1} with 
$\noise_t = {\cal O} ( \log t )$ and $\alpha = 0.5$. Moreover, $f(\prm)$ is strongly convex 
if $\lgp''( \theta ) \geq \mu$. For example, this holds for square loss function.   
Now if H\ref{iii3i} is also satisfied, repeating the analysis in Section~\ref{sec:main} 
yields an anytime and regret bound of ${\cal O} (\log t / t)$ and ${\cal O}( \log^2 T / T)$, respectively. 

We test our online MC algorithm on a small synthetically generated dataset, where $\bar{\prm}$ is a 
rank-20, $200 \times 5000$ matrix with Gaussian singular vectors. There are $2 \times 10^6$ observations
with Gaussian noise of variance $3$. 
Also, we test with two dataset \texttt{movielens100k}, \texttt{movielens20m} from \cite{movielens15}, which contains 
$10^5$, $2 \times 10^7$ movie ratings from $943$, $138493$ users on $1682$, $26744$ movies, respectively. 
We assume Gaussian observation and the loss function $f_t(\cdot)$ is designed as the 
square loss. 

\begin{figure}[t]
\centering
\includegraphics[height=.111\paperheight]{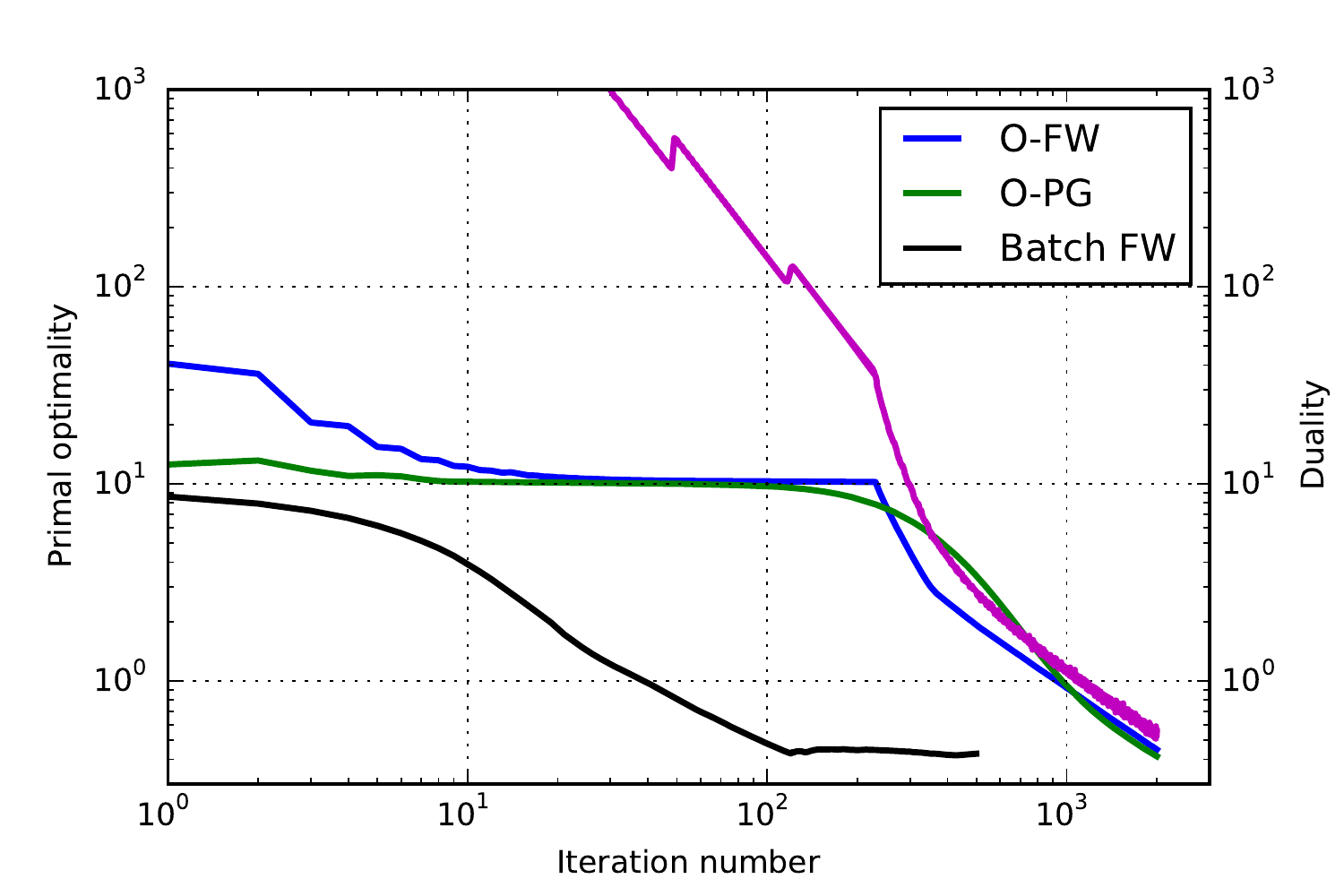}~\includegraphics[height=.111\paperheight]{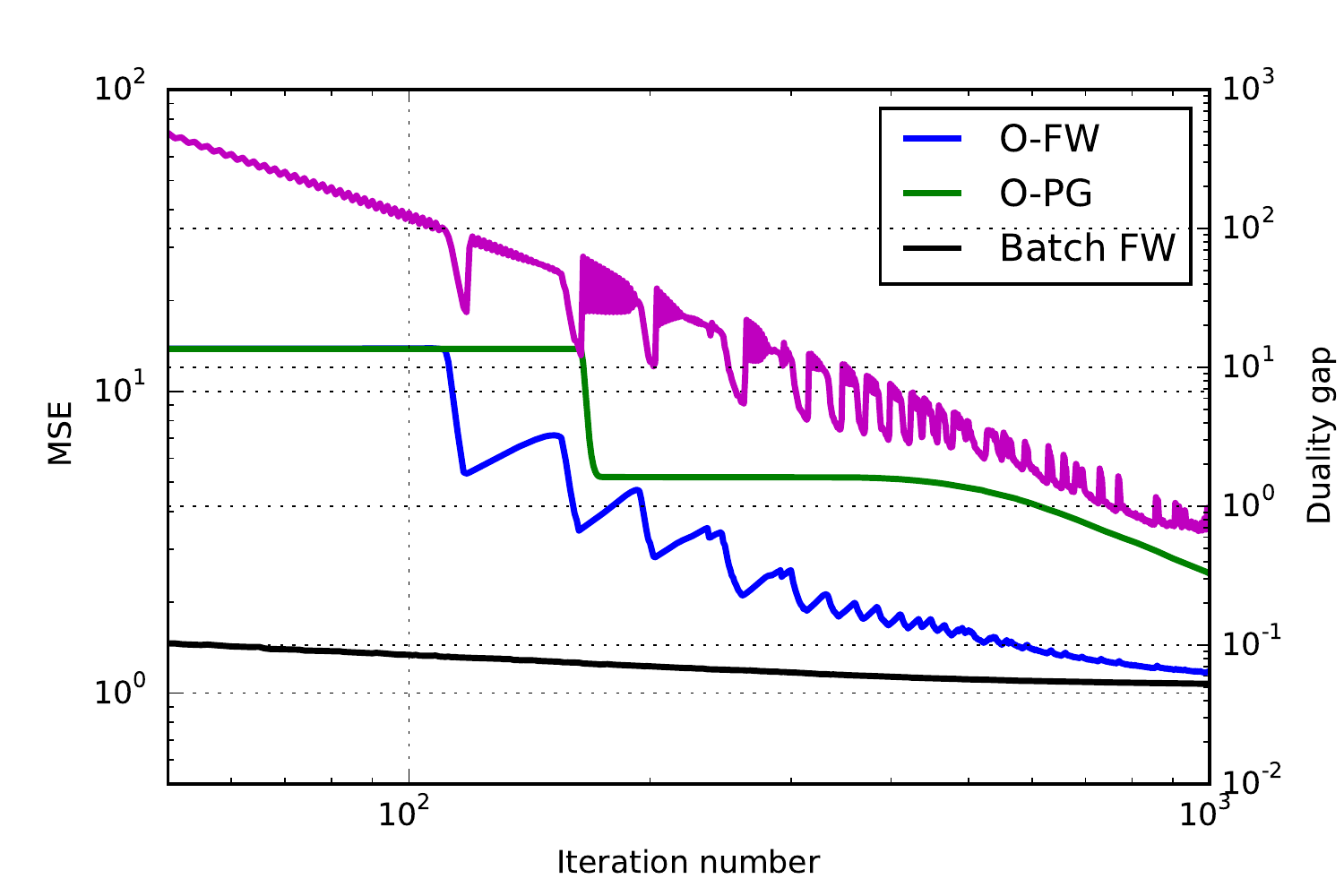}~\includegraphics[height=.111\paperheight]{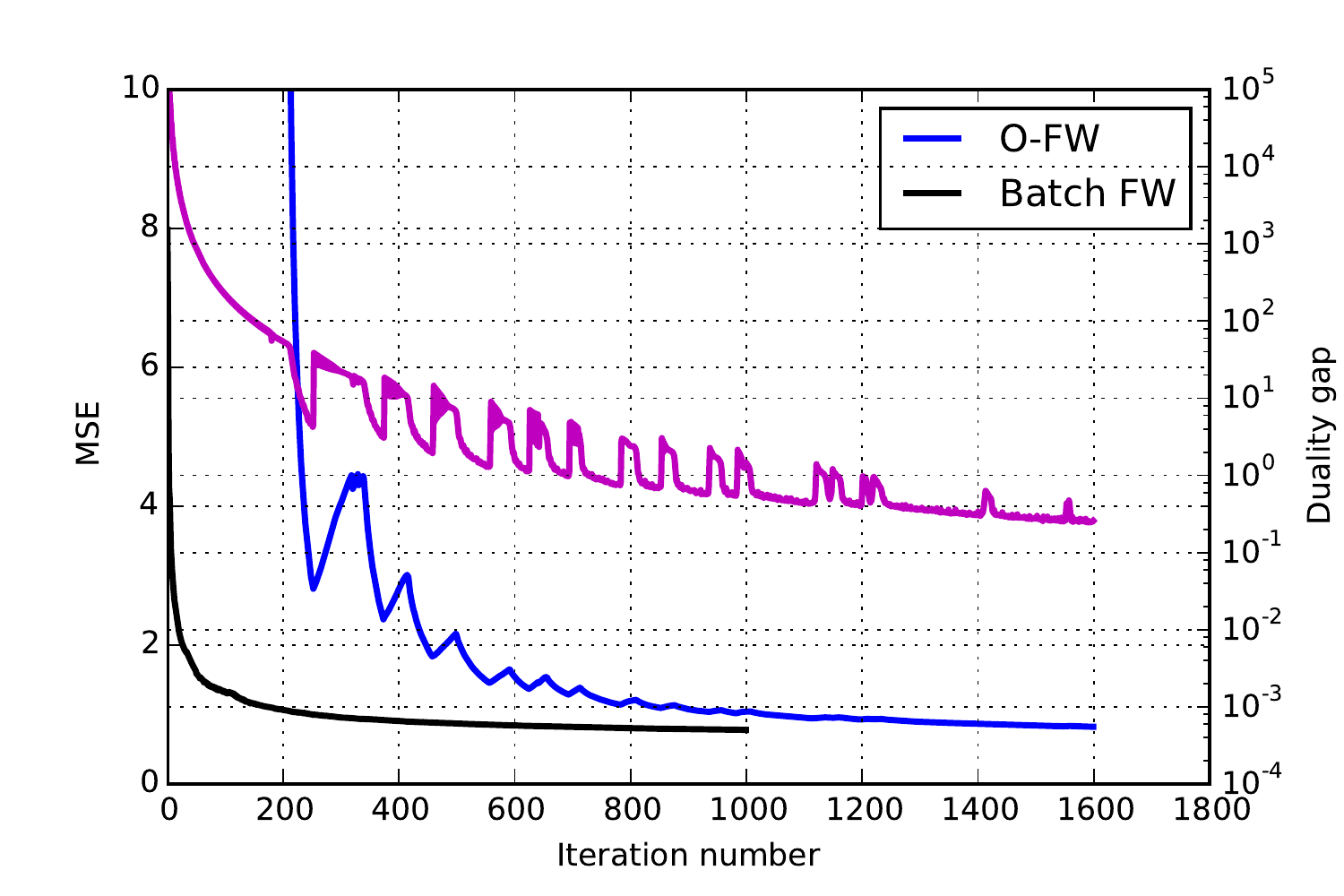} \\
\includegraphics[height=.111\paperheight]{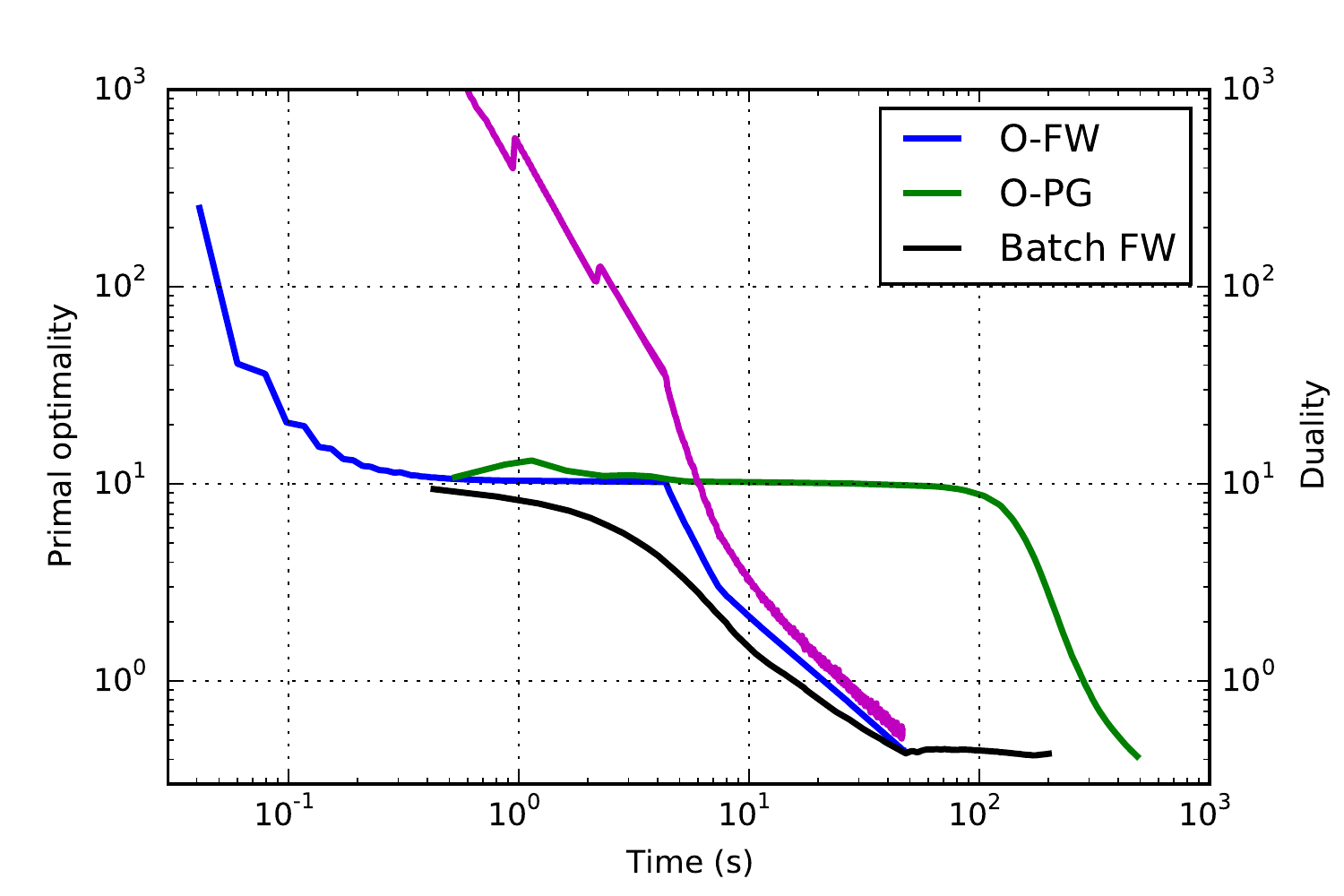}~\includegraphics[height=.111\paperheight]{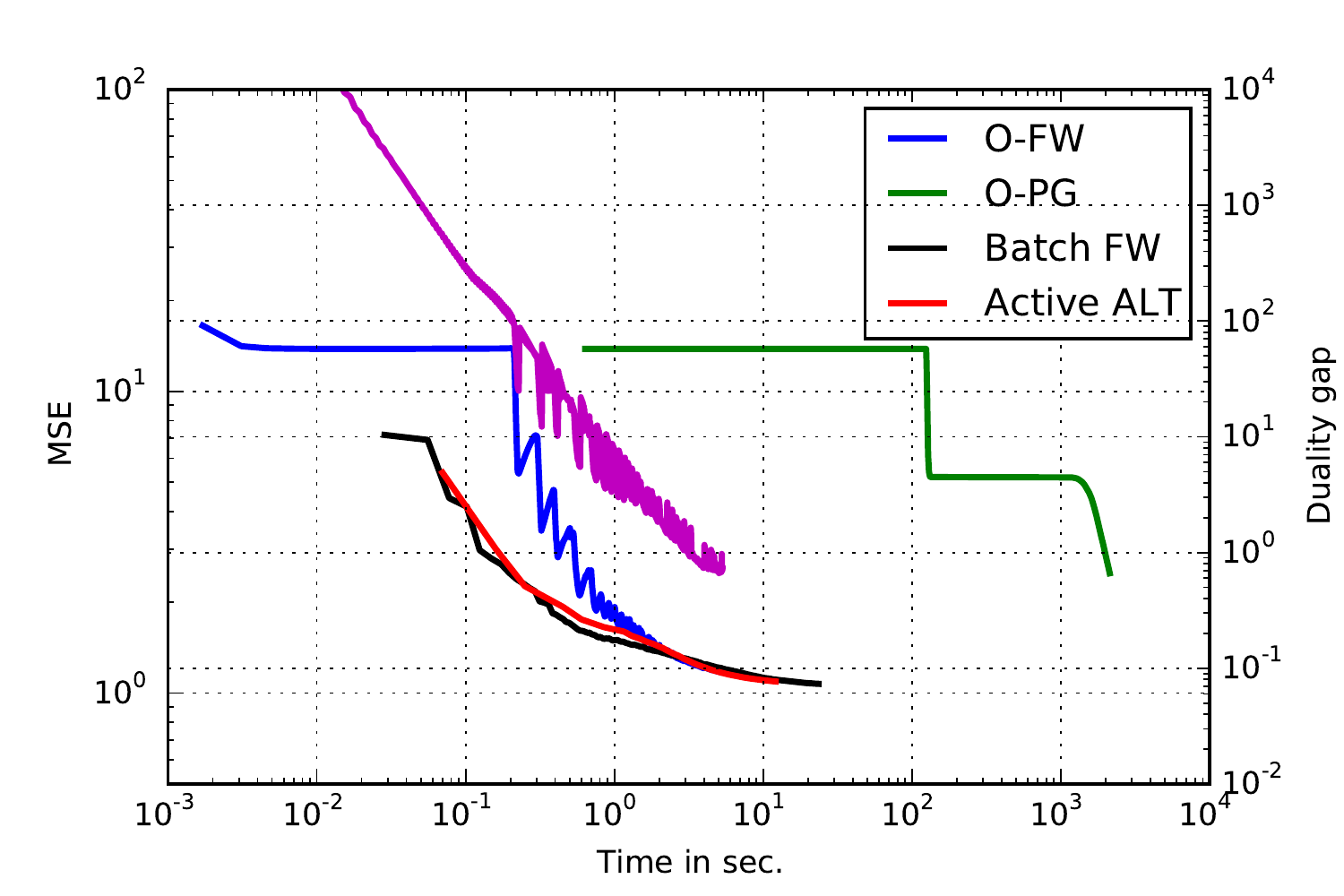}~\includegraphics[height=.111\paperheight]{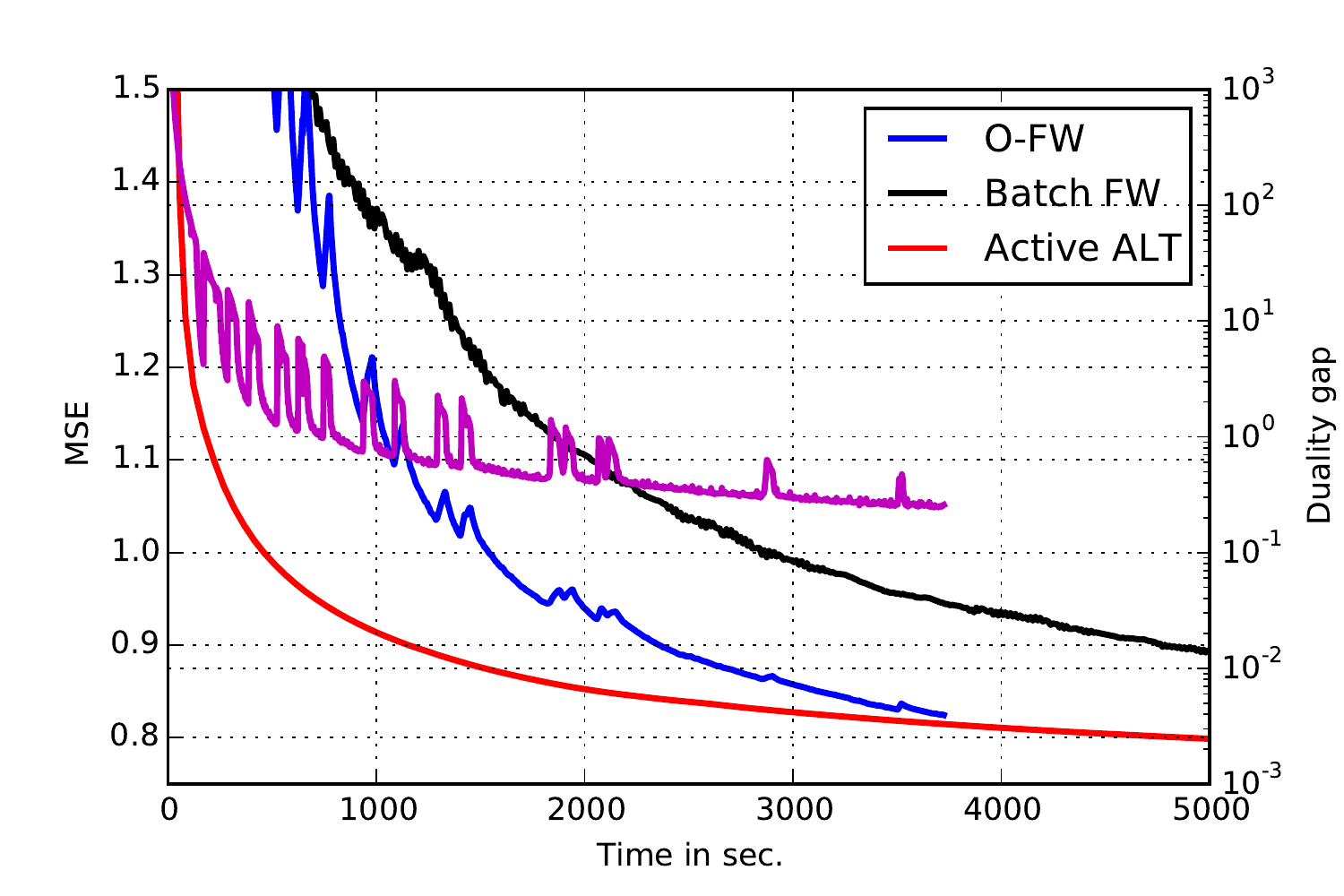} \vspace{-.2cm}
\caption{Online MC performance. (Left) synthetic with batch size $B=1000$; (Middle) \texttt{movielens100k} with $B=80$; (Right) \texttt{movielens20m} with $B=10000$. (Top) objective value/MSE against round number; (Bottom) against execution time. The duality gap $g_t^{\sf FW}$ for O-FW is plotted in purple.} \label{fig:mc} \vspace{-.2cm}
\end{figure}

\textbf{Results}. We compare O-FW 
to a simple online projected-gradient (O-PG) method. The step size for O-FW is set as
$\gamma_t = 2 / (1+t)$. For the movielens datasets, the parameter $\bar{\prm}$ is unknown,
therefore we split the dataset into training ($80\%$) and testing ($20\%$) set and evaluate
the mean square error on the test set. Radiuses of ${\cal C}_R$ are set as $R=1.1 \| \bar{\prm} \|_{\sigma,1}$ (synthetic), $R= 10000$ (\texttt{movielens100k}) and $R=150000$ (\texttt{movielens20m}). Note that
 H\ref{iii3i} is satisfied by the synthetic case. 

The results are shown in Figure~\ref{fig:mc}. For the synthetic data, 
we observe that the stochastic objective of O-FW decreases
at a rate $\sim {\cal O}(1/t)$, as predicted in our analysis. 
Significant complexity reduction compared to O-PG for synthetic and \texttt{movielens100k}
datasets are also observed. 
The running time is faster than the \emph{batch} FW with line searched step size 
on \texttt{movielens20m}, which we suspect is 
caused by the simpler linear optimization \eqref{eq:linear} 
solved at the algorithm initialization by 
O-FW\footnote{This operation amounts to finding the top singular vectors
of $\grd F_t(\prm_t)$, whose complexity grows linearly 
with the number of non-zeros in $\grd F_t( \prm_t)$.}; 
and is also comparable to a state-of-the-art,
specialized \emph{batch} algorithm for MC problems in \cite{hsieh14} (`active ALT') 
and achieves the same MSE level, 
even though the data are acquired in an online fashion in O-FW.  


\subsection{Example: Robust Binary Classification with Outliers} 

Consider the following online learning setting: the training data is given sequentially in the form of 
$(y_t, {\bm x}_t)$, where $y_t \in \{ \pm 1 \}$ is a binary label and ${\bm x}_t \in \RR^n$ 
is a feature vector. Our goal is 
to train a classifier $\prm \in \RR^{n}$ such that for an arbitrary 
feature vector $\hat{\bm x}$ it assigns $\hat{y} = {\rm sign} ( \langle \prm, \hat{\bm x} \rangle )$. 

\begin{figure}[t]
\centering
\includegraphics[height=.111\paperheight]{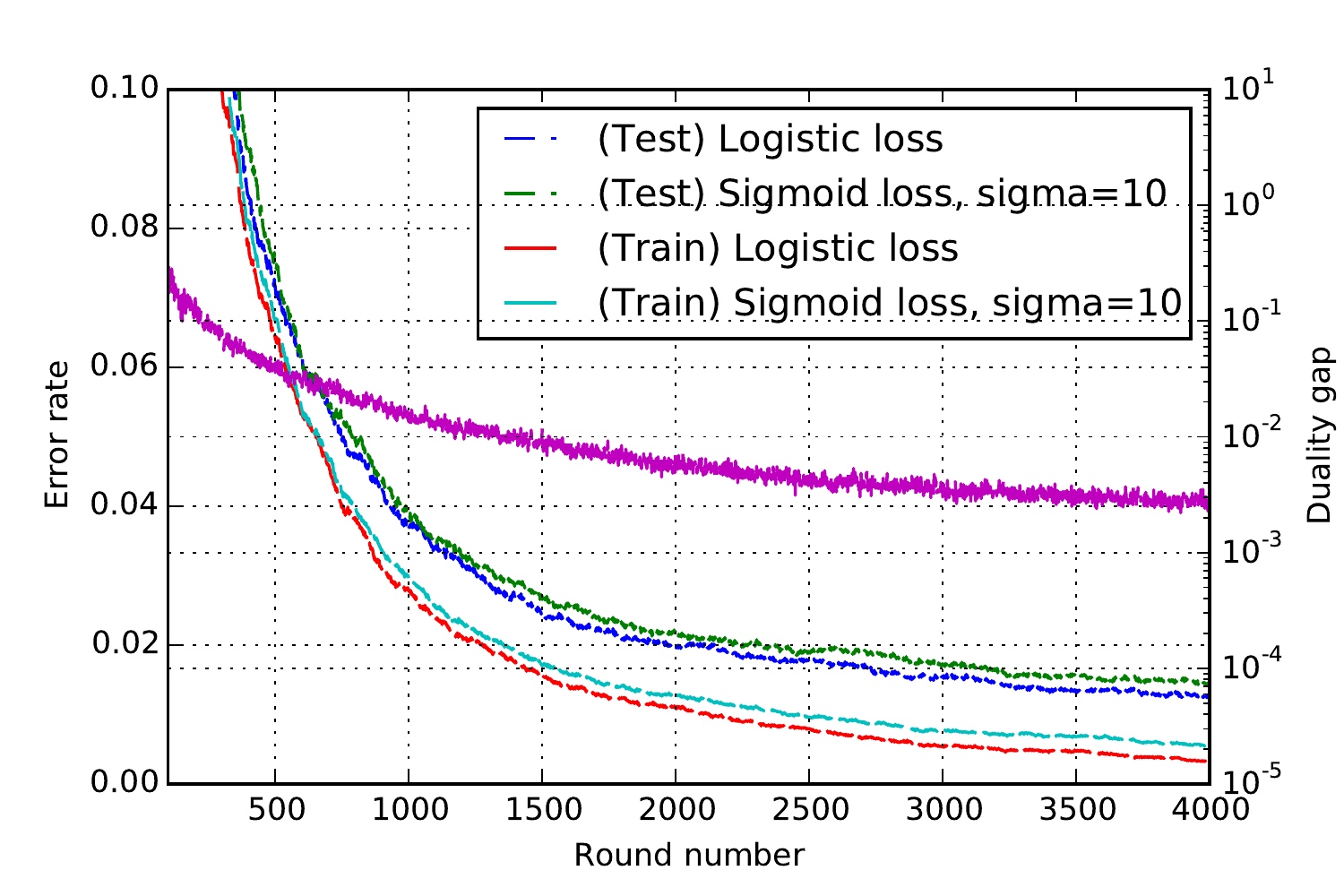}~\includegraphics[height=.111\paperheight]{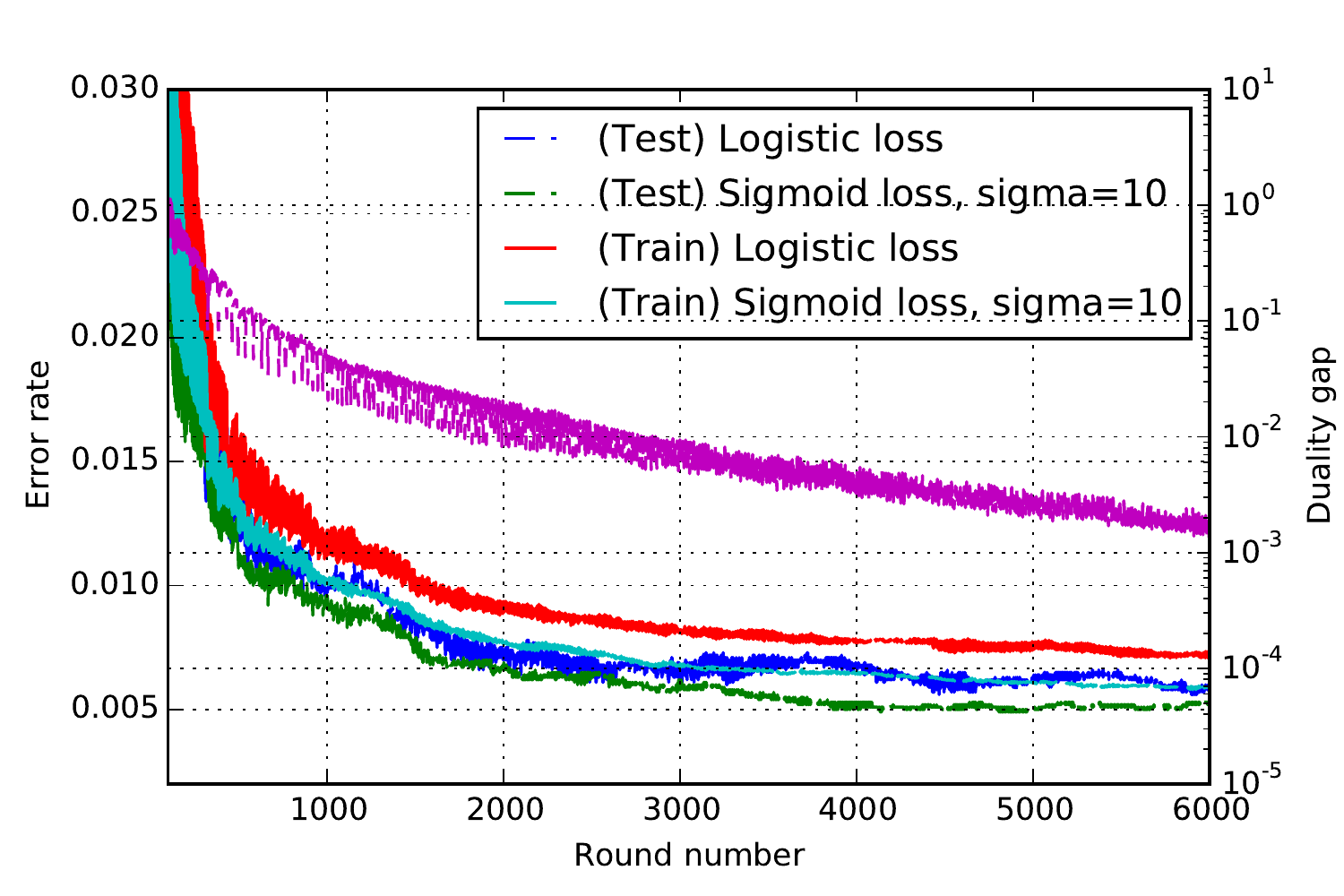}~\includegraphics[height=.111\paperheight]{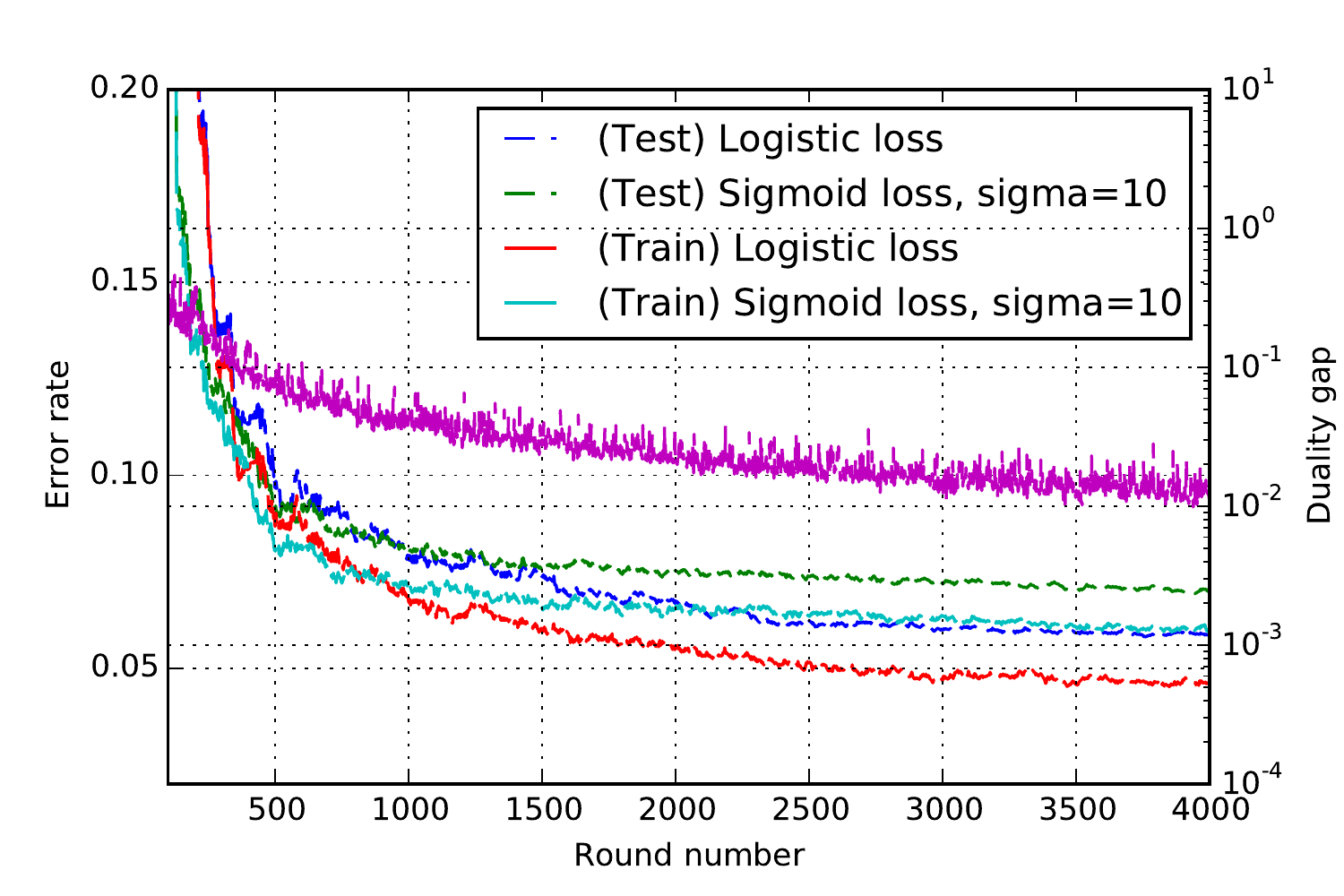} \\
\includegraphics[height=.111\paperheight]{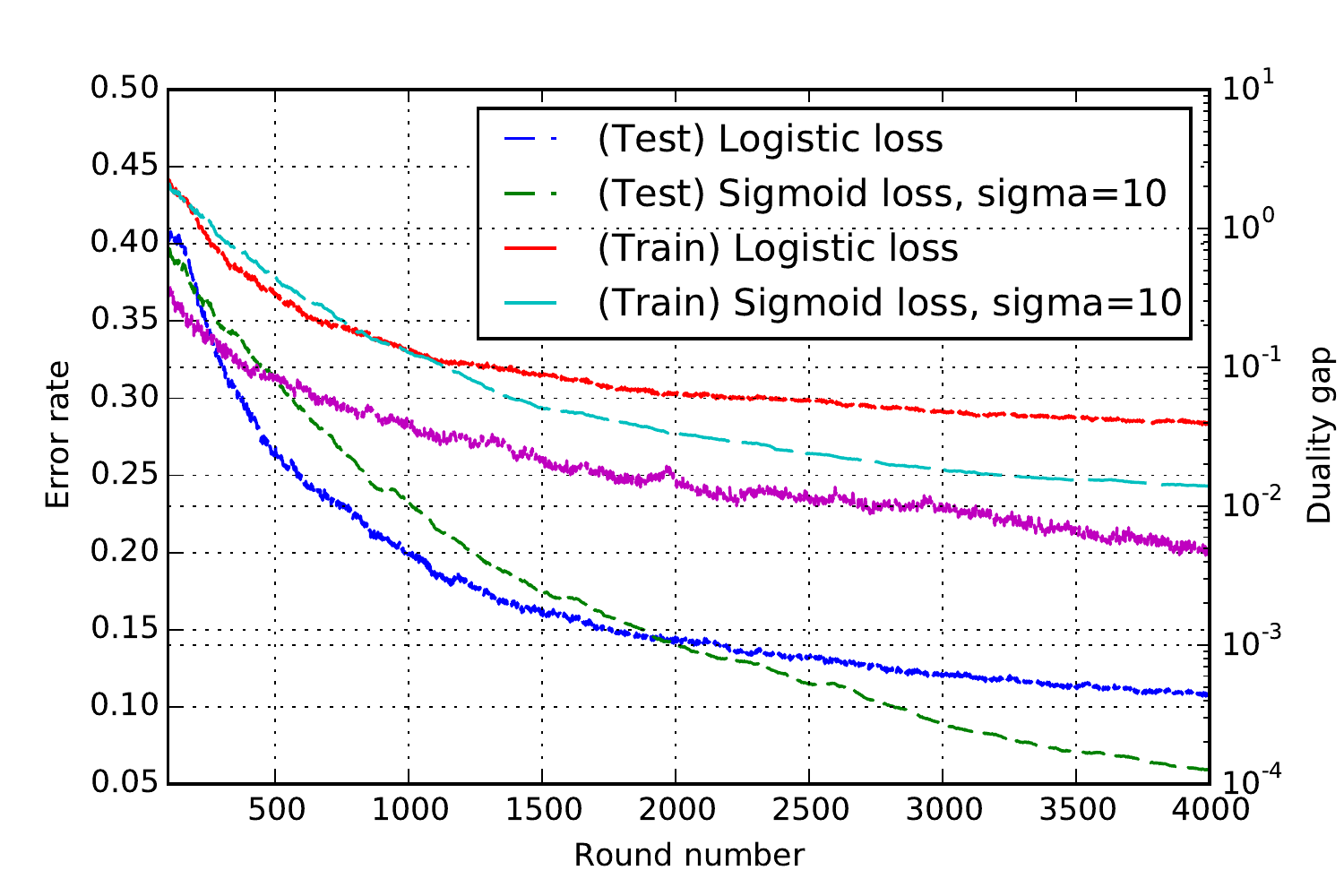}~\includegraphics[height=.111\paperheight]{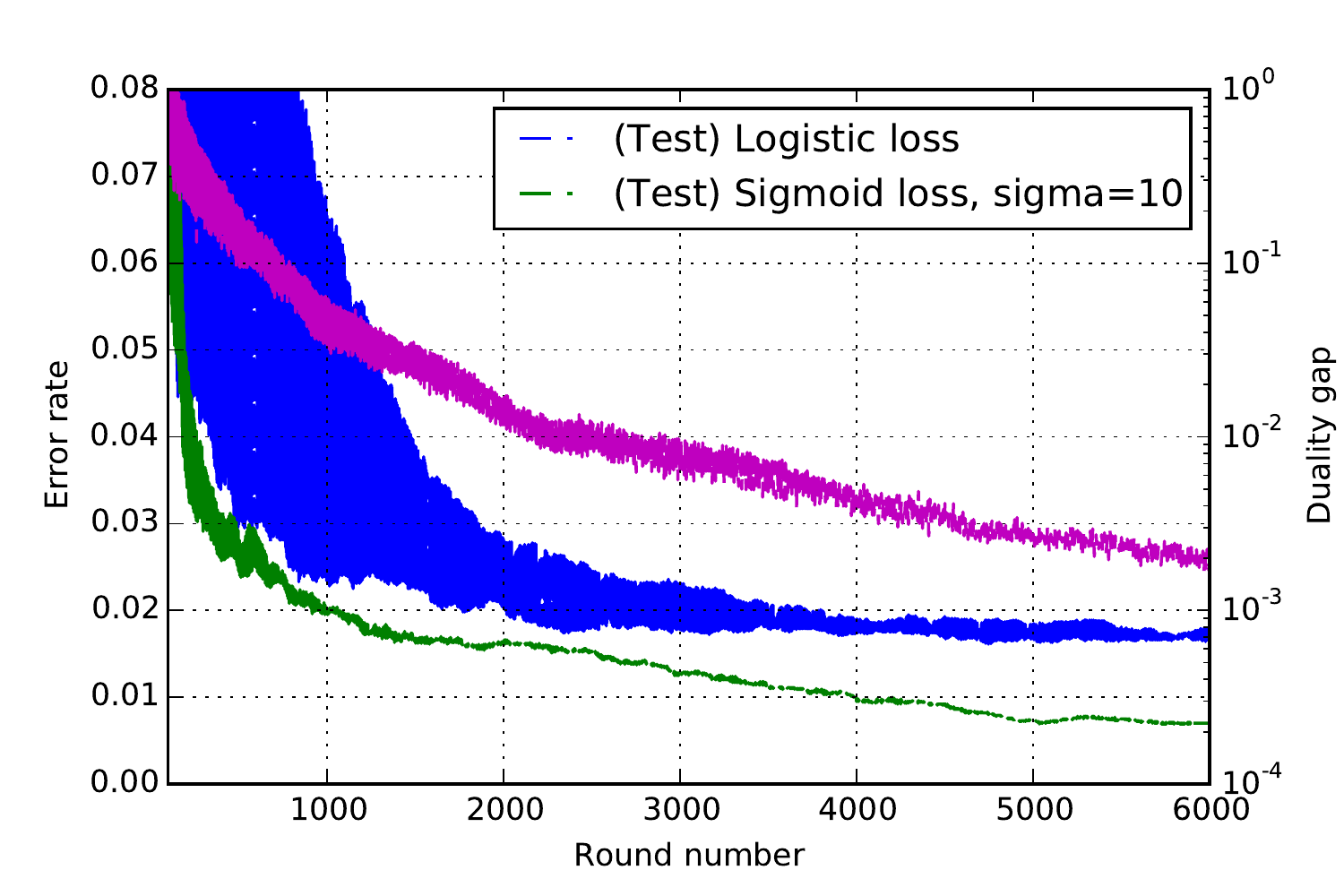}~\includegraphics[height=.111\paperheight]{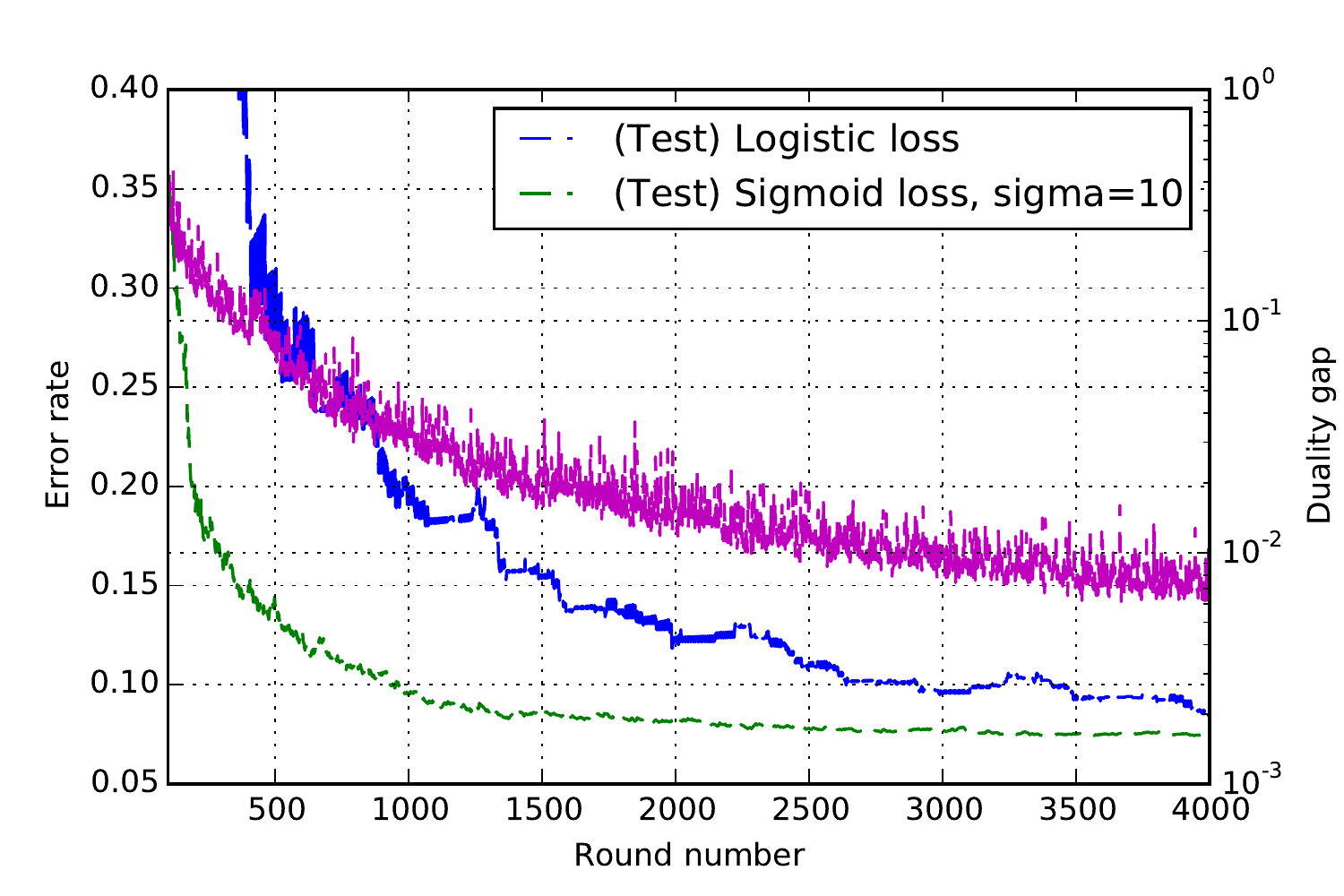} \vspace{-.4cm}
\caption{Binary classification performance against round number $t$ for: (Left) synthetic data; (Middle) \texttt{mnist} (class `1'); (Right) \texttt{rcv1.binary}. (Top) with no flip (Bottom) with $25\%$ flip in the training labels. The duality gap $g_t^{\sf FW}$ 
for O-FW with sigmoid loss is plotted in 
purple.} \label{fig:classifier} \vspace{-.2cm}
\end{figure}

The dataset may sometimes be contaminated by wrong labels. 
As a remedy, we design a \emph{sigmoid} loss function $f_t( \prm) 
\eqdef (1 + {\rm exp}( 10 \cdot y_t \langle \prm, {\bm x}_t \rangle ) )^{-1}$ that approximates the
$0/1$ loss function \cite{shalev11,bottou11}. 
Note that $f_t (\prm)$ is smooth and Lipschitz, but \emph{not convex}. 
For $\Cset$, we consider the $\ell_1$ ball $\Cset_{\ell_1} = 
\{ \prm \in \RR^n : \| \prm \|_1 \leq r \}$ when a sparse classifier is preferred; or the trace-norm 
ball $\Cset_\sigma = \{ \prm \in \RR^{m_1 \times m_2} : \| \prm \|_{\sigma,1} \leq R \}$, where $n = m_1 m_2$, 
when 
a low rank classifier is preferred. 

We evaluate the performance of our online classifier on synthetic and real data. 
For the synthetic data, the true classifier $\bar{\prm}$ is a rank-$10$, $30 \times 30$ Gaussian matrix. 
Each feature ${\bm x}_t$  is a 
$30\times 30$ Gaussian matrix. We have $40000$ ($20000$) tuples of data for training (testing).
We also test the classifier on the
\texttt{mnist} (classifying `1' from the rest of the digits), \texttt{rcv1.binary} dataset from \texttt{LIBSVM} \cite{CC01a}. 
The feature dimensions
are $784$, $47236$, and there are $60000$ ($10000$) and 
$20242$ ($677399$) data tuples for training (testing), respectively.
We artificially and randomly flip  
$0\%$, $25\%$ labels in the training set.

\textbf{Results}. 
As benchmark, we compare with
the logistic loss function, i.e., $f_t (\prm) = \log( 1 + {\rm exp} ( -y_t \langle \prm, {\bm x}_t \rangle ) )$.
We apply O-FW with a learning rate of $\alpha = 0.75$ for both loss functions,
i.e., $\gamma_t = 1/t^{0.75}$. 
For the synthetic data and \texttt{mnist}, the sigmoid 
(logistic) loss classifier is trained with a trace norm ball constraint of $R = 1$ ($R=10$). 
Each round is fed with a batch of $B=10$ tuples of data.
For \texttt{rcv1.binary}, we train the classifiers with $\ell_1$-ball constraint of $r=100$ ($r=1000$) 
for sigmoid (logistic) loss. Each round is fed with a batch of $B=5$ tuples of data.

As seen in \autoref{fig:classifier}, the logistic loss and sigmoid loss 
performs similarly when there are no flip in the labels; and the sigmoid loss demonstrates
better classification performance when some of the labels are flipped. Lastly, the duality gap of O-FW 
applied to the non-convex loss decays gradually with $t$, indicating that the algorithm
converges to a stationary point.





\appendix



\appendix

\section{Proof of Proposition~\ref{prop:ogrd}}
The following proof is an application of a modified version of \cite[Theorem 5]{sss09}\footnote{Note that
\cite[Theorem 5]{sss09} assumed implicitly that $\grd f( \prm; \omega_s )$ is bounded 
for all $\omega_s$, which can be generalized by our assumption that $\grd f( \prm; \omega_s )$ 
is sub-Gaussian.}. 
Let us define
\begin{equation}
\bm{\epsilon}_t (\prm)  = \grd F_t (\prm) - \grd f(\prm) = \frac{1}{t} \sum_{s=1}^t \Big( \grd \Obj (\prm; \omega_s) - \EE_{\omega \sim {\cal D}} [\grd \Obj (\prm; \omega) ] \Big) \eqs.
\end{equation}
From \cite{vg05}, for some sufficiently small $\epsilon > 0$, there exists a Euclidean $\epsilon$-net, $\mathcal{N}(\epsilon)$, 
with cardinality bounded by
\begin{equation}
 | \mathcal{N}(\epsilon)| = \mathcal{O}\left( n^2 \log(n)\left(\frac{\bar \rho }{ \epsilon}\right)^n\right) \eqs.
\end{equation}
In particular, for any $\prm \in \Cset$ there is a point $p_{\prm} \in \mathcal{N} (\epsilon / L)$ such that 
$\| p_{\prm} - \prm \|_2 \leq \epsilon / L$. This implies:
\begin{equation*}
\begin{split}
 \| \bm{\epsilon}_t (\prm) \|_\infty & \leq \|\bm{\epsilon}_t (p_{\prm})\|_\infty + \|\bm{\epsilon}_t (p_{\prm}) - \bm{\epsilon}_t (\prm)\|_\infty \leq \| \bm{\epsilon}_t (p_{\prm})\|_\infty + \|\grd F_t (\prm) - \grd F_t (p_{\prm})\|_\infty + \|\grd f (\prm) - \grd f (p_{\prm})\|_\infty \\
&\leq \| \bm{\epsilon}_t (p_{\prm})\|_\infty + \|\grd F_t (\prm) - \grd F_t (p_{\prm})\|_2 + \|\grd f (\prm) - \grd f (p_{\prm})\|_2 \leq \| \bm{\epsilon}_t (p_{\prm})\|_\infty + 2 L \| \prm - p_{\prm} \|_2 \\
 &\leq \|\bm{\epsilon}_t (p_{\prm})\|_\infty + 2\epsilon \eqs,
\end{split}
\end{equation*}
where we used the $L$-smoothness of $\grd F_t (\prm)$ and $\grd f (\prm)$ for the second last inequality.
Applying the union bound and controlling each point $p_{\prm} \in \mathcal{N} (\epsilon / L)$ 
using the sub-Gaussian assumption 
yields:
\begin{equation*}
\begin{split}
  \PP\Big( \sup_{ \prm \in \Cset } \| \bm{\epsilon}_t(\prm) \|_\infty > s \Big) \leq \PP \Big( \bigcup_{ p_{\prm} \in {\cal N}( \epsilon / L) } \Big\{ \|\bm{\epsilon}_t(p_{\prm}) \|_\infty > s - 2 \epsilon  \Big\} \Big) 
 &\leq |\mathcal{N}(\epsilon / L)| \cdot 2n \exp \left(-\frac{t(s - 2\epsilon)^2}{2 \sigma_D^2} \right) \\
& \hspace{-1.5cm} \leq \mathcal{O}\left( n^3 \log(n) \left(\frac{L \bar \rho }{ \epsilon} \right)^n \exp \left(-\frac{t(s - 2\epsilon)^2}{2 \sigma_D^2} \right)\right) \eqs. 
\end{split}
\end{equation*}
Setting $s=3\epsilon$ in the above, 
it can be verified that the following holds with probability at least $1 - \delta$
\begin{equation} \label{eq:largescale}
\| \bm{\epsilon}_t (\prm) \|_\infty = {\cal O} \left(  
\max\{ L \bar \rho , \:\: \sigma_D \} \sqrt{
\frac{ n \log(t) \log(n/\delta) }{t}} \right) \eqs.
\end{equation}
Applying another union bound over $t \geq 1$ (e.g., by setting $\delta = \epsilon / t^2$) 
then yields the desired result.

%
%
%


\section{Proof of Theorem~\ref{thm:sfw}} \label{sec:regretpf}
We define $h_t \eqdef f(\prm_t) - \min_{\prm \in \Cset} f(\prm)$ in the following. 
The analysis below is done by assuming a more general step size rule $\gamma_t = K / (K+t-1)$ 
with some $K \in \mathbb{Z}_+^\star$. 
First of all, we notice that for both \autoref{alg:ofw} and \autoref{alg:ofw_away} with the step size rule
$\gamma_t = K / (K+t-1)$, we have $\gamma_1 = 1$ and thus $h_1 = f(\atom_1) - f(\prm^\star)
 < \infty$.
For $t \geq 2$, we have the following convergence results for
 FW/AW algorithms with inexact gradients.

As explained in the proof sketch, let us state the following lemma which
is borrowed from \cite{simon_jaggi13,Lacoste_Jaggi15}.

\begin{Lemma} \cite{simon_jaggi13,Lacoste_Jaggi15}
\label{lem:stg_cvx}  \label{lem:ht_gtaw}
Assume H\ref{iii3i} and that $f$ is $L$-smooth and $\mu$-strongly convex, then
\begin{equation} \label{eq:gt_ht}
\big( \max_{\prm \in \Cset} \langle \grd f(\prm_t), \prm_t - \prm \rangle \big)^2 \geq 2 \mu \delta^2 {h_t} \quad {\rm and} \quad \Curve \geq \mu \delta^2 \eqs.
\end{equation}
Consider \autoref{alg:ofw_away}, assume H\ref{aw1} and that $f$ is $L$-smooth and $\mu$-strongly convex, then
\begin{equation} \label{eq:gtaw_ht}
 \big(\max_{ \prm \in {\cal A}_t } \langle \grd f(\prm_t), \prm \rangle - \min_{ \prm \in \Cset} \langle \grd f(\prm_t), \prm \rangle \big)^2 \geq 2 \mu \daw^2 h_t \quad {\rm and} \quad \Curve \geq \mu \daw^2 \eqs.
\end{equation}
\end{Lemma}
The above lemma is a key result that leads to
the linear convergence of
the classical FW/AW algorithms with \emph{adaptive} step sizes, as studied in \cite{simon_jaggi13,Lacoste_Jaggi15}.
Lemma~\ref{lem:ht_gtaw} enables us to prove the theorems below
for the FW/AW algorithms with inexact gradient and \emph{fixed} step sizes,
whose proof can be founded in \autoref{sec:pf_converge_fast} and \ref{sec:converge2}:


\begin{Theorem} \label{thm:converge_fast}
Consider Algorithm~\ref{alg:ofw} with the assumptions given in Theorem~\ref{thm:sfw}.
The following holds with probability at least $1-\epsilon$:
\begin{equation} \label{eq:bound}
\Obj(\prm_t) - \Obj(\prm^\star) \leq D_1 \left(\frac{\noise_t}{t+K-1}\right)^{2 \alpha},~\forall~t \geq 2 \eqs,
\end{equation}
where $\beta = 1 + 2 \alpha/(K-\alpha)$ and
\begin{equation*}
D_1 =
\max \Big\{4 \Big(\frac{K+1}{K}\Big)^{2\alpha}, \beta^2 \Big\} \cdot \frac{ (\rho \sigma + K \Curve/2)^2 }{ 2 \delta^2 \mu} \eqs.
\end{equation*}
\end{Theorem}

The anytime bound for \autoref{alg:ofw} is obvious from the above Theorem.

%

\begin{Theorem} \label{thm:converge2}
Consider Algorithm~\ref{alg:ofw_away} with the assumptions given in Theorem~\ref{thm:sfw}.
The following holds with probability at least $1-\epsilon$:
\begin{equation} \label{eq:bound2}
 \Obj(\prm_t) - \Obj(\prm^\star) \leq D_2 \Big( \frac{\noise_t}{\rstep{t-1}+K} \Big)^{2 \alpha},~\forall~t \geq 2, \eqs,
\end{equation}
where $n_t$ is the number of non-drop steps (see \autoref{alg:ofw_away}) up to iteration $t$,  $\beta = 1 + 2 \alpha / (K - \alpha)$ and
\begin{equation*}
D_2 =
\max \Big\{ \Big(\frac{K+1}{K}\Big)^{2\alpha}, \beta^2 \Big\} \cdot
\frac{2 ( 2\rho \sigma + K \Curve/2)^2 }{ (\daw)^2 \mu} \eqs.
\end{equation*}
\end{Theorem}
In addition, we have the following Lemma for \autoref{alg:ofw_away}.
\begin{Lemma}
\label{lem:drop_lb}
 Consider \autoref{alg:ofw_away}. We have
 $\rstep{t} \geq t/2$ for all $t$, where $\rstep{t}$ is the number of non-drop steps taken until round $t$.
\end{Lemma}
\begin{proof}
Except at initialization,
the active set is never empty.
Indeed, if there is only one active atom
left, then its weight is $1$.
Therfore the condition of line \ref{ln:drop_cond}
is satisfied and the atom cannot be dropped.
 Denote by $q_t$ the number of iterations
 where an atom was dropped up to time $t$ (line~\ref{ln:drop}).
 As noted above,
 $\rstep{t} + q_t =t$ holds. Since to be dropped,
 an atom needs to be added to the active set $\activ_t$
 first, $q_t \leq t/2$ also holds, yielding the result.
\end{proof}
Combining Theorem~\ref{thm:converge2} and the above lemma, we get the desirable
anytime bound for \autoref{alg:ofw_away}.

\subsection{Proof of Lemma~\ref{lem:stg_cvx}}
We first prove the first part of the lemma, i.e., \eqref{eq:gt_ht}, pertaining to the O-FW algorithm. 
Let $\bar{\bm s}_t \in \partial \Cset$ be a point on the boundary of $\Cset$ 
such that it is co-linear with $\prm^\star$ and $\prm_t$. 
Moreover, we defin
${g}_t \eqdef \max_{\prm \in \Cset} \langle \grd f(\prm_t), \prm_t - \prm \rangle$.
As $\prm^\star \in {\rm int}(\Cset)$, we can write
\beq
\prm^\star = \prm_t + \bar{\gamma} ( \bar{\bm s}_t - \prm_t)~~~\text{for some}~~~\bar{\gamma} \in [0,1) \eqs. 
\eeq
From the $\mu$-strong convexity of $f$, we have
\beq
\frac{\mu}{2} \| \prm^\star - \prm_t \|_2^2 \leq f ( \prm^\star ) - f( \prm_t ) - \langle \grd f( \prm_t ), \prm^\star - \prm_t \rangle = -h_t + \bar{\gamma} \langle \grd f( \prm_t ), \prm_t -  \bar{\bm s}_t \rangle \leq -h_t + \bar{\gamma} g_t \eqs,
\eeq
where the last inequality is due to the definition of $g_t$. Now, the left hand side of 
the inequality above can be bounded as
\beq
\frac{\mu}{2} \| \prm^\star - \prm_t \|_2^2 = \bar{\gamma}^2 \frac{\mu}{2} \| \bar{\bm s}_t - \prm_t \|_2^2 
\geq \bar{\gamma}^2 \frac{\mu}{2} \| \bar{\bm s}_t - \prm^\star \|_2^2 \geq \bar{\gamma}^2 \delta^2 \frac{\mu}{2}
\eeq
Combining the two inequalities above yields
\beq
h_t \leq \bar{\gamma} g_t - \bar{\gamma}^2 \delta^2 \frac{\mu}{2} \leq \frac{ g_t^2 }{ 2 \delta^2 \mu } \eqs,
\eeq
where the upper bound is achieved by setting $\bar{\gamma} = g_t / (\delta^2 \mu)$. 
Recalling the definition of $g_t$ concludes the proof of the first part. 
Lastly, 
we note by combining Eq.~(2), Remark 1 and Lemma 2 in \cite{simon_jaggi13},
we have $\Curve \geq \mu \delta^2$.


Next, we prove the second part of the lemma, i.e., \eqref{eq:gtaw_ht}, pertaining to the O-AW algorithm. 
Recall that as $\Cset$ is a polytope, we can write $\Cset = {\rm conv} ( {\cal A} )$ where ${\cal A}$ is a 
finite set of \emph{atoms} in $\RR^n$, i.e., $\Cset$ is a convex hull of ${\cal A}$. 
Note that ${\cal A}_t \subseteq {\cal A}$ for all $t$ in
the O-AW algorithm. 
Let us define the \emph{pyramidal width} $\daw$ of $\Cset$ as:
\begin{align} \label{eq:pywidth}
 &\daw \eqdef \inf_{\begin{subarray}{c}
{\cal K} \in {\rm faces}(\Cset),
\prm \in {\cal K},{\bm d} \in {\rm cone}( \Cset - \prm ) \setminus \{ 0 \} 
\end{subarray}} \inf_{ {\cal A}' \in {\cal A}_{\prm} } \frac{1}{\| {\bm d} \|_2} \Big( \max_{ {\bm y} \in {\cal A}' 
\cup \{ {\bm a}({\cal K}, {\bm d}) \} } \langle {\bm d}, {\bm y} \rangle 
- \min_{ {\bm y} \in {\cal A}' \cup \{ {\bm a}({\cal K}, {\bm d}) \} } \langle {\bm d}, {\bm y} \rangle \Big) \eqs,
\end{align}
where ${\cal A}_{\prm} \eqdef \{ {\cal A}' : {\cal A}' \subseteq {\cal A}~ 
\text{such that} ~\prm \in {\rm conv}({\cal A}')~\text{and $\prm$ is a proper convex combination of ${\cal A}'$}\}$
and ${\bm a}({\cal K}, {\bm d}) \eqdef \arg \max_{ {\bm v} \in {\cal K} } \langle {\bm v}, {\bm d} \rangle$.
Now, define the quantities:
\beq
\gamma^A ( \prm , \prm' ) \eqdef \frac{ \langle \grd f(\prm), \prm - \prm' \rangle }{ \langle \grd f(\prm), {\bm v}_f(\prm) - {\bm s}_f( \prm) \rangle} \eqs,
\eeq
where ${\bm v}_f(\prm) \eqdef \arg \min_{ \atom \in {\cal A}( \prm ) } \langle \grd f (\prm) , \atom \rangle$ and ${\bm s}_f ( \prm ) \eqdef \arg \min_{ \atom \in {\cal A} } \langle \grd f(\prm), \atom \rangle$. 
From \cite[Theorem 6]{Lacoste_Jaggi15}, it can be verified that
\beq \label{eq:ljjthm}
\mu \cdot \daw^2  \leq \inf_{ \prm \in \Cset }~\Big(  \inf_{ \prm' \in \Cset, {\rm s.t.} \langle \grd f(\prm), \prm' - \prm \rangle < 0 } \Big( \frac{2}{ \gamma^A( \prm, \prm' )^2 } \big( f(\prm') - f(\prm) - \langle \grd f(\prm), \prm' - \prm \rangle \big) \Big) \Big) \eqs,
\eeq
In the above, we have denoted ${\cal A}(\prm) \eqdef \{ {\bm v} = {\bm v}_{\cal A'} (\prm)~:~ {\cal A'} \in {\cal A}_{\prm} \}$ where ${\bm v}_{\cal A'} (\prm) \eqdef \arg \max_{ \atom \in {\cal A}' } \langle \grd f(\prm) , \atom \rangle$. 
We remark that ${\cal A} ( \prm_t ) \subseteq {\cal A}_t$. 
Note that $\gamma^A ( \prm , \prm' ) > 0$ as long as $\langle \grd f(\prm), \prm' - \prm \rangle < 0$ is satisfied. 

Assume $\prm_t \neq \prm^\star$ and observe that we have 
$\langle \grd f(\prm_t), \prm^\star - \prm_t \rangle < 0$, Eq.~\eqref{eq:ljjthm} implies that
\beq \label{eq:implyljj}
\frac{ \gamma^A ( \prm_t , \prm^\star )^2 }{2} \mu \daw^2 \leq f(\prm^\star) - f( \prm_t) -  \langle \grd f(\prm_t), \prm^\star - \prm_t \rangle = - h_t + \gamma^A ( \prm_t , \prm^\star ) \langle \grd f(\prm_t), {\bm v}_f(\prm_t) - {\bm s}_f( \prm_t) \rangle \eqs,
\eeq
where the equality is found using the definition of $ \gamma^A ( \prm_t , \prm^\star )$. 
Define $g_t^{AW} \eqdef \max_{ \prm \in {\cal A}_t } \langle \grd f(\prm_t), \prm \rangle - \min_{ \prm \in \Cset} \langle \grd f(\prm_t), \prm \rangle$ and
observe that 
\beq
\langle \grd f(\prm_t), {\bm s}_f (\prm_t) \rangle = \min_{ \prm \in \Cset } \langle \grd f(\prm_t), \prm \rangle~~~{\rm and}~~~\langle \grd f(\prm_t), {\bm v}_f (\prm_t) \rangle \leq \max_{ \prm \in {\cal A}_t }  \langle \grd f(\prm_t), \prm \rangle \eqs.
\eeq
Plugging the above into \eqref{eq:implyljj} yields
\beq
h_t \leq - \frac{ \gamma^A ( \prm_t , \prm^\star )^2 }{2} \mu \daw^2  +  \gamma^A ( \prm_t , \prm^\star ) g_t^{AW} \leq \frac{ (g_t^{AW})^2 }{ 2 \daw^2 \mu } \eqs,
\eeq
where we have set $\gamma^A ( \prm_t , \prm^\star ) = g_t^{AW} / ( \daw^2 \mu)$ similar to the first 
part of this proof. 
This concludes the proof for the lower bound on $g_t^{AW}$. 
Lastly, it follows from Remark~7, Eq.~(20) and Theorem~6 of \cite{Lacoste_Jaggi15} that
$\mu \daw^2 \leq \Curve$.

\section{Proof of Theorem~\ref{thm:dual}} \label{sec:oaw}
In the following, we denote the minimum loss action at round $t$ as $\prm_t^\star \in \arg
\min_{ \prm \in \Cset } F_t (\prm)$. Notice that $F_t(\prm)$ may be non-convex. 

Observe that for O-FW:
 \beq \begin{split}
 F_t( \prm_{t+1} ) & \leq F_t( \prm_t ) + \gamma_t \langle \grd F_t(\prm_t), \atom_t - \prm_t \rangle + \frac{1}{2} \gamma_t^2 \Curve 
 =  F_t( \prm_t ) - \gamma_t g_t^{\sf FW} + \frac{1}{2} \gamma_t^2 \Curve,
 \end{split}
 \eeq
 where the first inequality is due to the fact that $f$ is $L$-smooth and $\Cset$ has a diameter of $\bar{\rho}$.
 Define $\Delta_t \eqdef F_t(\prm_t) - F_t(\prm_t^\star)$ to be the
 instantaneous loss at round $t$ (recall that $\prm_t^\star \in \arg \min_{ \prm \in \Cset } F_t( \prm)$). 
 We have
 \beq \begin{split} \label{eq:ht}
 \Delta_{t+1} & = \frac{t}{t+1} \left( F_{t} (\prm_{t+1}) - F_{t} (\prm_{t+1}^\star) \right) + \frac{1}{t+1} ( f_{t+1} (\prm_{t+1}) - f_{t+1} (\prm_{t+1}^\star) ) \\
 \end{split}
 \eeq
 Note that the first part of the right hand side of \eqref{eq:ht} can be upper bounded as
 \beq \label{eq:firstpart}
 \begin{split}
 F_{t} (\prm_{t+1}) - F_{t} (\prm_{t+1}^\star) & \leq F_{t} (\prm_{t+1}) - F_{t} (\prm_{t}^\star) \leq \Delta_t - \gamma_t g_t^{\sf FW} + \frac{1}{2} \gamma_t^2 \Curve, \\
 \end{split}
 \eeq
 where the first inequality is due to $\prm_{t+1}^\star \in \Cset$ and the optimality of $\prm_t^\star$
 and the second inequality is due to the $L$-smoothness of $F_t$.
 Combining \eqref{eq:ht} and  \eqref{eq:firstpart} gives
 \[ \begin{split}
 \Delta_{t+1} & \leq \frac{t}{t+1} \left( \Delta_t - \gamma_t g_t^{\sf FW} + \gamma_t^2 \Curve / 2 \right) + \frac{1}{t+1} ( f_{t+1} (\prm_{t+1}) - f_{t+1} (\prm_{t+1}^\star) )  \\
 \Longleftrightarrow \frac{t}{t+1} \gamma_t g_t^{\sf FW} & \leq \frac{t}{t+1} \left( \Delta_t + \frac{1}{2} \gamma_t^2 \Curve \right) + \frac{1}{t+1} ( f_{t+1} (\prm_{t+1}) - f_{t+1} (\prm_{t+1}^\star) ) - \Delta_{t+1} \eqs.
 \end{split}
 \]
 Using the definition of $\Delta_{t+1}$, we note that $(t+1)^{-1} ( f_{t+1} (\prm_{t+1}) - f_{t+1} (\prm_{t+1}^\star) ) - \Delta_{t+1} = -(t/(t+1)) (F_t( \prm_{t+1} ) - F_t( \prm_{t+1}^\star) )$. Therefore, simplifying terms give
 \beq \label{eq:key_ineq}
 \gamma_t g_t^{\sf FW} \leq \Delta_t - ( F_t(\prm_{t+1}) - F_t(\prm_{t+1}^\star) ) + \gamma_t^2 \Curve / 2 \eqs.
 \eeq
 Observe that:
 \[ \begin{split}
 \sum_{t=T/2+1}^T \Big( \Delta_t - ( F_t(\prm_{t+1}) - F_t(\prm_{t+1}^\star) ) \Big) & = \sum_{t=T/2+1}^T \Big( ( F_t( \prm_t ) -  F_t(\prm_{t+1}) ) - ( F_t(\prm_t^\star) - F_t(\prm_{t+1}^\star) ) \Big) \vspace{.2cm} \\
 &\hspace{-5cm} = -F_T(\prm_{T+1}) + F_{T/2+1}(\prm_{T/2+1}) - F_{T/2+1}(\prm_{T/2+1}^\star) + F_T(\prm_{T+1}^\star) \\
 & \textstyle \hspace{-4.5cm} + \sum_{t=T/2+2}^T t^{-1} \Big( f_t(\prm_t) - F_{t-1} (\prm_t) - (f_t(\prm_t^\star) - F_{t-1} (\prm_t^\star)) \Big) \\
 &\hspace{-5cm} \textstyle \leq G \cdot \Big( \| \prm_{T+1} - \prm_{T+1}^\star \|_* + \| \prm_{T/2+1} - \prm_{T/2+1}^\star \|_* + \sum_{t=T/2+2}^T 2 t^{-1} \| \prm_t - \prm_t^\star \|_* \Big) \\
 &\hspace{-5cm} \textstyle \leq 2 \rho G \cdot \Big( 1 + \sum_{t=T/2+2}^T t^{-1} \Big) \leq 2 \rho G \cdot ( 1 + \log 2 ) \leq 4 \rho G \eqs.
 \end{split}
 \]
 where we have used the fact that $F_t(\prm_t) - F_{t-1}(\prm_t) = t^{-1} ( f_t(\prm_t) - F_{t-1} (\prm_t) )$
 in the first equality and that $f_t, F_t$ are $G$-Lipschitz in the second inequality.
 We notice that $\sum_{t=T/2+1}^T \gamma_t^2 = \sum_{t=T/2+1}^T  t^{-2 \alpha} \leq \log 2 \leq 1$ as $\alpha \in [0.5,1]$.
 Summing up both side of the inequality \eqref{eq:key_ineq} gives
 \beq
 \textstyle \Big( {\displaystyle \min_{t \in [T/2+1,T]} g_t^{\sf FW}} \Big) \cdot \sum_{t=T/2+1}^T \gamma_t
 \leq \sum_{t=T/2+1}^T \gamma_t g_t \leq 4 \rho G + \Curve / 2 \eqs,
 \eeq
 where the inequality to the left is due to $\gamma_t, g_t^{\sf FW} \geq 0$.
 Observe that for all $T \geq 6$,
 $
 \sum_{t=T/2+1}^T \gamma_t = \sum_{t=T/2+1}^T t^{-\alpha} \geq \frac{T^{1-\alpha} }{1 - \alpha} \left( 1 - \left( \frac{2}{3} \right)^{1-\alpha} \right) = \Omega \big( T^{1-\alpha} \big)$.
 We conclude that
 \beq
 \min_{t \in [T/2+1,T]} g_t^{\sf FW} \leq \frac{1-\alpha}{T^{1-\alpha}} (4 \rho G + \Curve / 2) \left( 1 - \left( \frac{2}{3} \right)^{1-\alpha} \right)^{-1} = {\cal O} ( 1 / T^{1 - \alpha} ) \eqs.\eeq

For the O-AW algorithm, we observe that
\beq
F_t( \prm_{t+1} ) \leq F_t ( \prm_t ) + \hat{\gamma}_t \langle \grd F_t( \prm_t ), {\bm d}_t \rangle + \frac{1}{2} \hat{\gamma}_t^2 \Curve
\eeq
Note that by construction, $\langle \grd F_t(\prm_t), {\bm d}_t \rangle = \min\{
\langle \grd F_t(\prm_t), \atom_t^{\sf FW} - \prm_t \rangle, \langle \grd F_t(\prm_t), \prm_t - \atom_t^{\sf AW}  \rangle \}$. Using the inequality $\min\{a,b\} \leq (1/2) (a+b)$, we have
\beq \label{eq:aw_sp}
F_t( \prm_{t+1} ) \leq F_t ( \prm_t ) + \hat{\gamma}_t \Big\langle \grd F_t( \prm_t ), \frac{1}{2} \Big( \atom_t^{\sf FW}  - \atom_t^{\sf AW} \Big) \Big\rangle + \frac{1}{2} \hat{\gamma}_t^2 \Curve
= F_t(\prm_t) - \frac{1}{2} \hat{\gamma}_t g_t^{\sf AW} + \frac{1}{2} \hat{\gamma}_t^2 \Curve.
\eeq
Proceeding in a similar manner to the proof for O-FW above, we get
\beq \label{eq:keyineq_aw}
\frac{1}{2} \hat{\gamma}_t g_t^{\sf AW} \leq \Delta_t - ( F_t(\prm_{t+1}) - F_t(\prm_{t+1}^\star) ) + \frac{1}{2} \hat{\gamma}_t^2 \Curve.
\eeq
The only difference from \eqref{eq:key_ineq} in the O-FW analysis
are the terms that depend on the actual step size $\hat{\gamma}_t$.

Now, Lemma~\ref{lem:drop_lb} implies that at least $T/4$ non-drop steps could have taken until round $T/2$,
therefore we have $\hat{\gamma}_t \leq \gamma_{T/4}$ for all $t \in [T/2+1,T]$ since
if a non-drop step is taken, then the step size will decrease; or if a drop-step step is taken, we have
$\hat{\gamma}_t \leq \gamma_{\rstep{t-1}}$ and $\rstep{t-1} \geq T/4$.
Therefore,
\[
\frac{1}{2} \sum_{t=T/2+1}^T  \hat{\gamma}_t^2 \Curve \leq \frac{T}{4} \cdot \Curve \Big( \frac{T}{4} \Big)^{-2 \alpha} \leq \Curve.
\]
Summing the right hand side of \eqref{eq:keyineq_aw} from $t=T/2+1$ to $t=T$ yields an upper bound
of $4 \rho G + \Curve$.

On the other hand, define
${\cal T}_{\text{non-drop}}$ be a subset of $[T/2+1,T]$ where a
non-drop step is taken.
We have
\[
\sum_{t=T/2+1}^T \hat{\gamma}_t \geq \sum_{ t \in {\cal T}_{\text{non-drop}} } \gamma_{\rstep{t}} \geq
\sum_{t=3T/4+1}^{T} \gamma_t \geq \frac{T^{1-\alpha}}{1-\alpha} \Big( 1 - \Big( \frac{4}{5} \Big)^{1-\alpha} \Big) = \Omega( T^{1-\alpha} ),
\]
where the second inequality is due to the fact that
$|{\cal T}_{\text{non-drop}}| \geq T/4$ and
the last inequality holds for all $T \geq 20$.
Finally, summing the left hand side of \eqref{eq:keyineq_aw} from $t=T/2+1$ to $t=T$ yields
\[
\Big( \min_{t \in [T/2+1,T]} g_t^{\sf AW} \Big) \cdot \sum_{t=T/2+1}^T \hat{\gamma}_t \leq
\sum_{t=T/2+1}^T \hat{\gamma}_t g_t^{\sf AW} \leq 4 \rho G + \Curve.
\]
Therefore, we conclude that $\min_{t \in [T/2+1,T]} g_t^{\sf AW} = {\cal O} ( 1 / T^{1-\alpha} )$ for the O-AW algorithm.

\section{Proof of Proposition~\ref{cor:ncvx}}
We first look at the O-FW algorithm. Our goal is to bound the following inner product
\[
\max_{ \prm \in \Cset } \langle \grd f( \prm_t), \prm_t - \prm \rangle,
\]
where $t \in [T/2+1,T]$ is the round index that satisfies $g_t^{\sf FW} = {\cal O} (1 / T^{1-\alpha})$, which exists due to Theorem~\ref{thm:dual}. 
For all $\prm \in \Cset$, observe that 
\beq \label{eq:ncvx_fw} \begin{split}
\langle \grd f( \prm_t), \prm_t - \prm \rangle & \leq \langle \grd F_t( \prm_t), \prm_t - \prm \rangle + \langle  \grd f(\prm_t) -  \grd F_t( \prm_t), \prm_t - \prm \rangle \\
& \leq g_t^{\sf FW} + \rho \| \grd f(\prm_t) -  \grd F_t( \prm_t) \|. \end{split}
\eeq
Following the same line of analysis as Proposition~\ref{prop:ogrd}, with probability at least $1-\epsilon$, 
it holds that
\beq \label{eq:ncvx_prob}
\| \grd f(\prm_t) -  \grd F_t( \prm_t) \|_\infty = {\cal O} \Big( \max\{ \sigma_D , \bar\rho L \} \sqrt{ \frac{n \log(t) \log(n /\epsilon) }{t}}  \Big),
\eeq
which is obtained from \eqref{eq:largescale}. 
Note that compared to Proposition~\autoref{prop:ogrd}, we save a factor of $\log(t)$ inside the square root as
the iteration instance $t$ is fixed. 
Using the fact that $t \geq T/2+1$, the following holds with probability at least $1-\epsilon$, 
\[
\langle \grd f( \prm_t), \prm_t - \prm \rangle = {\cal O} \big( \max \big\{ 1 / T^{1-\alpha} , \sqrt{ \log T  / T } \big\} \big),~\forall~ \prm \in \Cset.
\]

For the O-AW algorithm, we observe that the inequality \eqref{eq:aw_sp} in \autoref{sec:oaw} can be replaced by
\[
F_t( \prm_{t+1} ) \leq F_t(\prm_t) - \hat{\gamma}_t \langle \grd F_t(\prm_t), \prm_t - \atom_t^{\sf FW} \rangle + \frac{1}{2} \hat{\gamma}_t^2 \Curve.
\]
Furthermore, we can show that the inner product $\langle \grd F_t(\prm_t), \prm_t - \atom_t^{\sf FW} \rangle$
decays at the rate of ${\cal O}(1/T^{1-\alpha})$ by replacing $g_t^{\sf AW}$ in the proof in Appendix~\ref{sec:oaw} with this inner product. 
Consequently, \eqref{eq:ncvx_fw} holds for the $\prm_t$ generated by O-AW, i.e.,
\[
\langle \grd f( \prm_t), \prm_t - \prm \rangle \leq \langle \grd F_t(\prm_t), \prm_t - \atom_t^{\sf FW} \rangle + \rho \| \grd f(\prm_t) -  \grd F_t( \prm_t) \|
\]
Applying \eqref{eq:ncvx_prob} yields our result.

\section{Proof of Theorem~\ref{thm:converge_fast}}
\label{sec:pf_converge_fast}
This section establishes a ${\cal O} ( ( \noise_t / (t+K-1) )^{2\alpha} )$ bound for $h_t$
for \autoref{alg:ofw} with inexact gradients, i.e., replacing $\grd F_t(\prm_t)$ by $\Hgrd_t f (\prm_t)$
satisfying H\ref{iii1}, under the assumption that $f(\prm)$ is $L$-smooth, $\mu$-strongly convex and $\gamma_t = K / (K+t-1)$.

Define $\bm{\epsilon}_t = \Hgrd_t f(\prm_t) - \grd f(\prm_t) $,
$g_t = \max_{ {\bm s} \in \Cset } \langle  \prm_t - {\bm s} , \nabla \Obj({\bm \theta}_t) \rangle$
as the duality gap at $\prm_t$.
Notice that \eqref{eq:gt_ht} in  Lemma~\ref{lem:stg_cvx} implies:
\beq \label{eq:gt_s}
{g}_t \geq \sqrt{ 2\mu \delta^2 h_t},
\eeq
Define ${\bm s}_t \in \argmax_{ {\bm s} \in \Cset } \langle  \prm_t - {\bm s} , \nabla \Obj({\bm \theta}_t) \rangle$.
We note that
\begin{align}
\langle \nabla f(\prm_t) , {\atom}_t - \prm_t \rangle
& \leq \langle \hat{\nabla} \Obj (\prm_t) , {\bm s}_t - \prm_t \rangle
- \langle \bm{\epsilon}_t , {\atom}_t - \prm_t \rangle = \langle \nabla \Obj (\prm_t) , {\bm s}_t - \prm_t \rangle
+  \langle \bm{\epsilon}_t , {\bm s}_t - \atom_t \rangle \nonumber\\
& \leq - g_t +  \rho \| \bm{\epsilon}_t \| \leq - \delta \sqrt{2 \mu h_t} +  \rho \| \bm{\epsilon}_t \|\eqs,\label{eq:chain2K}
\end{align}
where the last line follows from \eqref{eq:gt_s}.
Combining the $L$-smoothness of $f(\prm)$ and \eqref{eq:chain2K} yield
the following with probability at least $1- \epsilon$ and for all $t \geq 1$,
\begin{equation}
 \label{eq:2ndeqK}
h_{t+1} \leq \sqrt{h_t} (\sqrt{h_t} - \gamma_t\delta \sqrt{2 \mu})
+ \gamma_t\rho \sigma \left( \frac{\noise_t}{t+K-1} \right)^\alpha + \frac{1}{2} \gamma_t^2 \Curve \eqs.
\end{equation}
Let us recall the definition of $D_1$
\begin{equation}
D_1 =
\max\{4 ((K+1)/K)^{2\alpha}, \beta^2 \} ( \rho \sigma + K \Curve/2)^2 /( 2 \delta^2 \mu)
\quad \text{with} \quad \beta = 1 + 2 \alpha/(K-\alpha) \eqs,
\end{equation}

and proceed by induction. Suppose that $h_t \leq D_1 ( \noise_t / (t+K-1))^{2\alpha}$ for some
$t \geq 1$. There are two cases.\\
\textbf{Case 1} $h_t - \gamma_t\delta \sqrt{2 \mu h_t} \leq 0$:\\
Then since $\gamma_t = K/(K+t-1)$, \eqref{eq:2ndeqK} yields
\begin{align}
\label{eq:case_negK}
 h_{t+1} &\leq \rho \sigma K\frac{({\noise_t})^\alpha}{(K+t-1)^{1 + \alpha}} + \frac{\Curve K^2}{2 (K+t-1)^2} \leq (\rho \sigma K + \Curve K^2 /2) \frac{(\noise_{t+1})^{2\alpha}}{(K+t-1)^{2\alpha}} \nonumber \\
 & \leq (\rho \sigma K + \Curve K^2 /2) \Big( \frac{K+1}{K}\Big)^{2\alpha} \Big( \frac{\noise_{t+1}}{K+t} \Big)^{2\alpha} \eqs, \nonumber
\end{align}
where we used that $\noise_t$ is increasing and larger than $1$.
To conclude, one just needs to check that
\begin{equation}
 (\rho \sigma K + \Curve K^2 /2) \Big( \frac{K+1}{K}\Big)^{2\alpha} \leq D_1 \eqs.
\end{equation}
Note that we have
\begin{equation}
D_1 \geq \Big( \frac{K+1}{K}\Big)^{2\alpha} (\rho \sigma + \Curve K /2) \cdot 4 \frac{ (\rho \sigma + \Curve K /2) }{2 \mu \delta^2} \geq \Big( \frac{K+1}{K}\Big)^{2\alpha} (\rho \sigma + \Curve K /2) \cdot K \eqs,
\end{equation}
where the last inequality is due to $\Curve \geq \delta^2 \mu$ from Lemma~\ref{lem:stg_cvx}.
Hence,
\begin{equation*}
 h_{t+1}  \leq D_1 (\noise_{t+1} /(K+t))^{2\alpha} \eqs.
\end{equation*}

\textbf{Case 2} $h_t - \gamma_t\delta \sqrt{2 \mu h_t} > 0$:\\
By induction hypothesis and \eqref{eq:2ndeqK}, we have
 \begin{align}
 \label{eq:chain3K}
&h_{t+1} - D_1 \Big( \frac{\noise_{t+1}}{K+t} \Big)^{2\alpha} \nonumber \\
&\leq D_1 ( \Big( \frac{\noise_t}{K+t-1} \Big)^{2\alpha} - \Big( \frac{\noise_{t+1}}{K+t} \Big)^{2\alpha} )
+ \frac{ (\noise_t)^\alpha \cdot K }{ (K+t-1)^{1+\alpha} } \Big(
\rho \sigma + \Curve K / 2 -  \delta \sqrt{2 \mu D_1 } \Big) \nonumber \\
&\leq \frac{ (\noise_t)^{\alpha}    }{(K+t-1)^{1+ \alpha} }
\left[ 2\alpha D_1 \left(\frac{\noise_t}{t + K-1}\right)^\alpha
+K \rho \sigma + K^2 \Curve / 2 - \delta K \sqrt{2 \mu D_1 }
\right] \nonumber \\
&\leq \frac{ (\noise_t)^{\alpha}    }{(K+t-1)^{1+ \alpha} }
\left[ 2\alpha D_1 \left(\frac{\noise_t}{t + K-1}\right)^\alpha
+(K \rho \sigma + K^2 \Curve / 2)(1-\beta)\right]
\end{align}
where we used the fact that (i) $\noise_t$ is
increasing and larger than $1$, (ii) $t \geq 1$ and
(iii) $1/(K+t-1)^{2\alpha} - 1/(K+t)^{2\alpha} \leq 2 \alpha / (K+t-1)^{1+2\alpha}$ in the second last inequality;
and we have used the definition of $D_1$ in the last inequality.
Define
\begin{equation}
t_0 \eqdef \inf \{t  \geq 1 : 2\alpha D_1 \left(\frac{\noise_t}{t + K-1}\right)^\alpha
+(K \rho \sigma + K^2 \Curve / 2)(1-\beta) \leq 0 \}.
\end{equation}
Since $\noise_t /(K+ t-1)$ is monotonically decreasing to $0$ and $\beta > 1$, $t_0$ exists.
Clearly, for any $t > t_0$  the RHS is non-positive. For
$t\leq t_0$, we have
\begin{align}
\label{eq:rec2}
(K \rho \sigma + K^2 \Curve / 2)(\beta -1) \leq 2\alpha D_1 \left(\frac{\noise_t}{t + K-1}\right)^\alpha
\end{align}
\ie
\begin{align}
\label{eq:rec2}
D_0(K-\alpha)(\beta -1) \leq 2\alpha D_1 \left(\frac{\noise_t}{t + K-1}\right)^\alpha
\end{align}
Hence by the definition that $ \beta = 1 + 2 \alpha/(K-\alpha)$
and applying Theorem~\ref{thm:converge} (see Section~\ref{sec:pf_converge}) we get:
\begin{equation*}
 h_t \leq  D_0\left(\frac{\noise_t}{t + K-1}\right)^{\alpha}\leq D_1 \left(\frac{\noise_t}{t + K-1}\right)^{2\alpha}
\end{equation*}
The initialization is easily verified as the first inequality holds true for all $t \geq 2$.

\subsection{Proof of Theorem~\ref{thm:converge}}
\label{sec:pf_converge}

\begin{Theorem} \label{thm:converge}
Consider Algorithm~\ref{alg:ofw} and
assume H\ref{iii1} and that $f(\prm)$ is convex and $L$-smooth.
Then, the following holds with probability at least $1 - \epsilon$:
\begin{equation} \label{eq:bound1}
\Obj(\prm_t) - \Obj(\prm^\star) \leq  D_0 \left(\frac{\noise_t}{{t+K-1}}\right)^\alpha,~\forall~t \geq 2 \eqs,
\end{equation}
where
\begin{equation}
 \label{eq:def_D}
D_0 = \frac{K^2 \Curve / 2 + \rho \sigma K}{ K-\alpha} \eqs.
\end{equation}
\end{Theorem}

Let us define $h_t = \Obj(\prm_t) - \Obj(\prm^\star)$, then we get
\begin{equation}
\label{eq:baseqK}
h_{t+1} \leq h_t + \gamma_t \langle \nabla f(\prm_t) , {\atom}_t - \prm_t \rangle
+ \frac{1}{2} \gamma_t^2 \Curve \eqs.
\end{equation}
On the other hand, the following also hods:
\begin{align}
\langle \nabla f(\prm_t) , {\atom}_t - \prm_t \rangle
& =  \langle \hat{\nabla} \Obj (\prm_t)
 , {\atom}_t - \prm_t \rangle - \langle \bm{\epsilon}_t , {\atom}_t - \prm_t \rangle,  \nonumber \\
& \leq \langle \hat{\nabla} \Obj (\prm_t) , \prm^\star - \prm_t \rangle
- \langle \bm{\epsilon}_t , {\atom}_t - \prm_t \rangle  \nonumber \\
& = \langle \nabla \Obj (\prm_t) , \prm^\star - \prm_t \rangle
+  \langle \bm{\epsilon}_t , \prm^\star - \atom_t \rangle \nonumber\\
& \leq - h_t +  \rho \| \bm{\epsilon}_t \|\eqs. \label{eq:chain1K}
\end{align}
where the second line follows from the definition of $\atom_t$ and
the last inequality is due to the convexity of $\Obj$
and the definition of the diameter.
Plugging \eqref{eq:chain1K} into \eqref{eq:baseqK}
and using H\ref{iii1} yields the following with probability at least $1- \Delta$ and for all $t \geq 1$
\begin{equation} \label{eq:1steqK}
h_{t+1} \leq (1 - \gamma_t)h_t
+ \gamma_t\rho \sigma (\frac{\noise_t}{K+t-1})^\alpha + \frac{1}{2} \gamma_t^2 \Curve \eqs.
\end{equation}
We now proceed by induction to prove the first bound of the Theorem.
Define
$$D_0 = (K^2 \Curve / 2 + \rho \sigma K) / (K-\alpha) \eqs.$$
The initialization is done by applying \eqref{eq:1steqK} with $t=1$ and noting that $K \geq 1$.
Assume that $h_t \leq D_0 (\noise_t/(K+t-1))^\alpha$ for some $t \geq 1$.
Since $\gamma_t = K/(t+K-1)$,
from \eqref{eq:1steqK} we get:
\begin{align}
 &h_{t+1} - D_0 \Big( {\frac{\noise_{t+1}}{ K+t}} \Big)^\alpha\\
 &\leq D_0 \Big( \big( \frac{\noise_t}{t+K-1} \big)^\alpha - \big( \frac{\noise_{t+1}}{t+K} \big)^\alpha \Big) + \frac{K^2 \Curve / 2 + \rho \sigma K (\noise_t)^\alpha - D_0 K (\noise_t)^\alpha}{(t+K-1)^{1+\alpha}} \nonumber\\
 & \leq (\noise_t)^\alpha \Big( \frac{D_0}{(t+K-1)^\alpha} - \frac{D_0}{(t+K)^\alpha} + \frac{K^2 \Curve / 2 + \rho \sigma K - D_0 K}{(t+K-1)^{1+\alpha}} \Big) \nonumber\\
 &\leq \frac{(\noise_t)^\alpha}{(t+K-1)^{1+\alpha}} \Big( (\alpha -K) D_0 + K^2 \Curve / 2 + \rho \sigma K \Big) \leq 0 \eqs, \nonumber
\end{align}
where we used  the fact that $\noise_t$ is increasing
 and larger that $1$ for the second inequality
 and $1/(t+K-1)^\alpha - 1/(t+K)^\alpha \leq \alpha / (t+K-1)^{1+\alpha}$ for the third inequality.
 The induction argument is now completed.


\section{Proof of Theorem \ref{thm:converge2}} \label{sec:converge2}
This section establishes a ${\cal O} ( ( \noise_t / (\rstep{t-1} + K) )^{2\alpha} )$ bound for $h_t$ 
for \autoref{alg:ofw_away} with inexact gradients, i.e., replacing $\grd F_t(\prm_t)$ by $\Hgrd_t f (\prm_t)$ 
satisfying H\ref{iii1}, under the assumption that $f(\prm)$ is $L$-smooth, $\mu$-strongly convex and $\gamma_t = K / (K+t-1)$. 

\textbf{Outline of the proof.} Here, our strategy parallels that of 
\autoref{sec:pf_converge_fast}. We first 
show that the 
slow convergence rate of ${\cal O} ( (\noise_t / (\rstep{t-1} + K))^{\alpha} )$ holds for 
\autoref{alg:ofw_away} (Theorem \ref{thm:aw_slow}). 
The fast convergence rate of ${\cal O} ( (\noise_t / (\rstep{t-1} + K) )^{2\alpha} )$ is then established 
using induction. We have to pay special attention to the case when a drop step
is taken (line 13 of \autoref{alg:ofw_away}). In particular, when a drop step is taken, the induction step is done by 
Lemma \ref{lem:drop}; for otherwise, we apply similar arguments in 
\autoref{sec:pf_converge_fast} to proceed with the induction. 

To begin our proof, let us define $\bm{\epsilon}_t = \Hgrd_t f(\prm_t) - \grd f(\prm_t) $,
\begin{equation*}
\btom_t^{\sf FW} \eqdef 
\arg \min_{ {\btom} \in \Cset} ~\big\langle {\btom}, {\nabla} \Obj({\prm}_t) \big\rangle,~
\btom_t^{\sf AW} \eqdef \arg \max_{ {\btom} \in \activ_t } ~\big\langle {\btom},
{\nabla} \Obj({\prm}_t) \big\rangle,~
\bar{g}_t^{\sf AW}  \eqdef \pscal{\nabla \Obj( \prm_t)}{\btom_t^{\sf AW} - \btom_t^{\sf FW}} \eqs. 
\end{equation*}
We remark that ${\bm b}_t^{\sf AW} \neq \atom_t^{\sf AW}$ and ${\bm b}_t^{\sf FW} \neq \atom_t^{\sf FW}$
as they are evaluated on the true gradient $\grd f(\prm_t)$. 

Recall that in \autoref{alg:ofw_away}, we choose ${\bm d}_t$ such that $\pscal{ \Hgrd \Obj(\prm_t)}{ {\bm d}_t} = \min\{ \pscal{ \Hgrd \Obj(\prm_t)}{ \atom_t^{\sf FW} - \prm_t }, \pscal{ \Hgrd \Obj(\prm_t)}{ \prm_t - \atom_t^{\sf AW} } \}$. Therefore, 
for $t\geq2$:
\begin{align*}
 \pscal{ \Hgrd \Obj(\prm_t)}{ {\bm d}_t}& \leq \pscal{\Hgrd \Obj(\prm_t) }{ \frac{\atom_t^{\sf FW} - \atom_t^{\sf AW}}{2}} 
 \leq \pscal{\Hgrd \Obj(\prm_t) }{ \frac{\btom_t^{\sf FW} - \btom_t^{\sf AW}}{2}} \\
 &=\pscal{\grd \Obj(\prm_t) }{ \frac{\btom_t^{\sf FW} - \btom_t^{\sf AW}}{2}} 
 + \pscal{\bm{\epsilon}_t }{ \frac{\btom_t^{\sf FW} - \btom_t^{\sf AW}}{2}}
\end{align*}
where the second inequality is due to the definitions of $\atom_t^{\sf FW}$ and $\atom_t^{\sf AW}$ in \eqref{eq:away}. 
Hence:
\begin{equation}
\label{eq:aw_rel}  
   \pscal{ \Hgrd \Obj(\prm_t)}{ {\bm d}_t} \leq - \frac{\bar{g}_t^{\sf AW}}{2} 
  + \pscal{\bm{\epsilon}_t }{ \frac{\btom_t^{\sf FW} - \btom_t^{\sf AW}}{2}} 
\end{equation}

As $f$ is $L$-smooth, the following holds,
\begin{align}
&\Obj(\prm_{t+1}) \leq \Obj (\prm_t)   + \hat{\gamma}_t \pscal{ \grd \Obj(\prm_t)}{ {\bm d}_t } + \frac{\Curve}{2}  \hat{\gamma}_t^2 \label{eq:taylor}\\
& = \Obj (\prm_t)  + \hat{\gamma}_t ( \pscal{ \Hgrd \Obj(\prm_t)}{ {\bm d}_t} - \pscal{ \bm{\epsilon}_t}{ {\bm d}_t })
+ \hat{\gamma}_t^2 \frac{\Curve}{2} \nonumber \\
& \leq \Obj (\prm_t)  -\hat{\gamma}_t \frac{ \bar{g}_t^{\sf AW}}{2}  
+ \hat{\gamma}_t \pscal{\bm{\epsilon}_t}{ \frac{ \btom_t^{\sf FW} - \btom_t^{\sf AW} }{2} - {\bm d}_t }
+ \hat{\gamma}_t^2 \frac{\Curve}{2} \nonumber
\end{align}
where we used \eqref{eq:aw_rel} for the last line.
Subtracting $\Obj(\prm^*)$ on both sides and applying H\ref{iii1}
yield
\begin{equation}
 \label{eq:dyn_aw_gen}
h_{t+1} \leq h_t  -\hat{\gamma}_t \frac{ \bar{g}_t^{\sf AW}}{2}  
+ 2 \hat{\gamma}_t\rho \sigma \left(\frac{\noise_t}{K+t-1}\right)^\alpha
+ \hat{\gamma}_t^2 \frac{\Curve}{2} \eqs,
\end{equation}
where we have used $\| (\btom_t^{\sf FW} - \btom_t^{\sf AW})/2 - {\bm d}_t \|_* \leq 2 \rho$.

We first establish a slow convergence rate of O-AW algorithm.
Define
\begin{equation}
\label{eq:def_dprime}
 D_2' = \frac{K}{ K-\alpha} (K \Curve/2  + 2\rho \sigma) \eqs.
\end{equation}
\label{sec:pf_converge2}
\begin{Theorem}
\label{thm:aw_slow}
Consider Algorithm~\ref{alg:ofw_away}.
Assume H\ref{iii1} and that $f(\prm)$ is convex and $L$-smooth,
the following holds with probablity
$1 - \epsilon$:
\begin{equation}
h_t \eqdef \Obj(\prm_t) - \Obj(\prm^\star) \leq D_2' \Big( \frac{\noise_t}{\rstep{t-1}+K} \Big)^{\alpha}  \eqs,
\end{equation}
for all $t \geq 2$. 
Here $D_2'$ is given in \eqref{eq:def_dprime}.
\end{Theorem}
\begin{proof}
See \autoref{sec:pf_aw_slow}.
\end{proof}

Let us recall the definition of $D_2$
\begin{equation*}
D_2 = 2
\max\{((K+1)/K)^{2\alpha}, \beta^2 \} ( 2 \rho \sigma + K \Curve/2)^2 /( \daw^2 \mu) 
\quad \text{with} \quad \beta = 1 + 2 \alpha/(K-\alpha) \eqs.
\end{equation*}

To prove Theorem \ref{thm:converge2}, we proceed by induction and assume
that for some $t \geq 2$,
$h_t \leq D_2 (\noise_t / (K+\rstep{t-1}))^{2\alpha}$
holds.
Notice that \eqref{eq:gtaw_ht} in Lemma~\ref{lem:ht_gtaw} gives:
\beq \label{eq:gtaw_s}
\bar{g}_t^{\sf AW} \geq \sqrt{ 2\mu \daw^2 h_t},
\eeq
Now, suppose that $h_t >0$ ($h_t = 0$ is discussed at the end of the proof).
Combining \eqref{eq:dyn_aw_gen} and \eqref{eq:gtaw_s} gives:
\begin{equation}
\label{eq:dyn}
h_{t+1} \leq h_t - \hat{\gamma}_t \daw \sqrt{ \frac{\mu  h_t}{2}}
+ 2 \hat{\gamma}_t  \rho \sigma \Big( \frac{ \noise_t}{\rstep{t-1}+K} \Big)^\alpha
+ \hat{\gamma}_t^2 \frac{\Curve}{2} \eqs.
\end{equation}
We have used the fact that $t - 1\geq \rstep{t-1}$. 

Consider two different cases. If a drop step is taken at iteration $t+1$, the induction step can be done by the following:
\begin{Lemma}
\label{lem:drop}
Suppose that $h_t \leq D_2 (\noise_t / (K+\rstep{t-1}))^{2\alpha}$ 
and that a drop step is taken at iteration $t+1$ (see Algorithm \ref{alg:ofw_away} line \ref{ln:drop}), then
\begin{equation}
 h_{t+1} \leq D_2  \left( \frac{\noise_{t+1}}{K+\rstep{t}} \right)^{2\alpha} \eqs,
\end{equation}
note that $\rstep{t} = \rstep{t-1}$ when a drop step is taken.
\end{Lemma}
\begin{proof}
 See \autoref{sec:pf_drop}.\vspace{-.2cm}
\end{proof}
The above lemma shows that the objective value does not increase when a drop step is taken. 

On the other hand, when a drop step is \emph{not} taken at iteration $t+1$, then from \autoref{alg:ofw_away}, 
we have $\hat{\gamma}_{t} = \gamma_{\rstep{t}} =  K/(K + \rstep{t} - 1)$
and $\rstep{t} = \rstep{t-1}+1$.
We consider the following two cases:

\textbf{Case 1:} If $h_t - \hat{\gamma}_t \daw \sqrt{ \frac{\mu  h_t}{2}}\leq 0$.\\
Then, since $\hat{\gamma}_t = K/(K+\rstep{t}-1)$ and $\rstep{t} \leq t$, \eqref{eq:dyn} yields
\begin{align}
\label{eq:case_negK}
 h_{t+1} &\leq 2 \rho \sigma K\frac{({\noise_t})^\alpha}{(K+\rstep{t}-1)^{1 + \alpha}} + \frac{\Curve K^2}{2 (K+\rstep{t}-1)^2}\\
 &\leq (2 \rho \sigma K + \Curve K^2 /2) \frac{(\noise_{t+1})^{2 \alpha}}{(K+\rstep{t}-1)^{2\alpha}} \nonumber \\
 & \leq (2 \rho \sigma K + \Curve K^2 /2) \Big( \frac{K+1}{K}\Big)^{2\alpha}  \left(\frac{\noise_{t+1}}{K+\rstep{t}}\right)^{2\alpha} \eqs, \nonumber
\end{align}
where we used that $\noise_t$ is increasing and larger than $1$.
To conclude, one just needs to check that
\begin{equation}
 (2 \rho \sigma K + \Curve K^2 /2) \Big( \frac{K+1}{K}\Big)^{2\alpha} \leq D_2 \eqs.
\end{equation}
Note that we have 
\begin{align}
D_2 & \geq \Big( \frac{K+1}{K}\Big)^{2\alpha} (2 \rho \sigma + \Curve K / 2 ) 
\cdot 2 \frac{ (2 \rho \sigma + \Curve K / 2 ) }{\mu \daw^2} \geq \Big( \frac{K+1}{K}\Big)^{2\alpha} (2 \rho \sigma + \Curve K / 2 ) \cdot K \eqs, \nonumber
\end{align}
where the last inequality is due to $\Curve \geq \daw^2 \mu$ from Lemma \ref{lem:ht_gtaw}. 
Hence,
\begin{equation*}
 h_{t+1}  \leq D_2 (\noise_{t+1} /(K+\rstep{t}))^{2\alpha} \eqs.
\end{equation*}

\textbf{Case 2:} Assume $h_t - \hat{\gamma}_t \daw \sqrt{ \frac{\mu  h_t}{2}} > 0$.\\
By induction and \eqref{eq:dyn}, we have
 \begin{align}
 \label{eq:chain3K}
h_{t+1} - & D_2 \Big( \frac{\noise_{t+1}}{K+\rstep{t}} \Big)^{2\alpha} 
 \nonumber \\
& \leq D_2 ( \Big( \frac{\noise_t}{K+\rstep{t}-1} \Big)^{2\alpha} - \Big( \frac{\noise_{t+1}}{K+\rstep{t}} \Big)^{2\alpha} ) 
+ \frac{ (\noise_t)^\alpha \cdot K }{ (\rstep{t}+K-1)^{1+\alpha} } \Big( 
2 \rho \sigma + C_f K / 2 -  \daw \sqrt{\frac{ \mu D_2}{2} } \Big)
\nonumber \\
&\leq \frac{ (\noise_t)^{\alpha}    }{(K+\rstep{t}-1)^{1+ \alpha} }
\big[ 2\alpha D_2 \left(\frac{\noise_t}{ \rstep{t}+K-1}\right)^\alpha
+ 2 K \rho \sigma + K^2 \Curve / 2 - \daw K \sqrt{\frac{ \mu D_2}{2} }
\big] \nonumber \\ 
&\leq \frac{ (\noise_t)^{\alpha}    }{(K+\rstep{t}-1)^{1+ \alpha} }
\big[ 2\alpha D_2 \left(\frac{\noise_t}{\rstep{t} + K-1}\right)^\alpha
 + (2K \rho \sigma + K^2 \Curve / 2)(1-\beta)\big]
\end{align}
where we used the fact that (i) $\noise_t$ is 
increasing and larger than $1$, (ii) $t \geq 1$ and 
(iii) $1/(K+t-1)^{2\alpha} - 1/(K+t)^{2\alpha} \leq 2 \alpha / (K+t-1)^{1+2\alpha}$ in the second last inequality; and we have used the definition of $D_2$ in the last inequality.
Define
\begin{equation}
 t_0 \eqdef \inf \{t  \geq 1 : 
 2\alpha D_2 \left(\frac{\noise_t}{\rstep{t} + K-1}\right)^\alpha
+K (2 \rho \sigma + K \Curve / 2)(1-\beta) \leq 0 \}. 
\end{equation}

Since $\noise_t /(K+ \rstep{t}-1)$ decreases
 to $0$ (see H\ref{iii1} and Lemma \ref{lem:drop_lb}), $t_0$ exists. 
Clearly, for any $t > t_0$  the RHS is non-positive. 
For
$t\leq t_0$, we have
\begin{align}
\label{eq:rec2}
K (2 \rho \sigma + K \Curve / 2)(\beta -1) \leq 2\alpha D_2 \left(\frac{\noise_t}{\rstep{t} + K-1}\right)^\alpha
\end{align}
implying
\begin{align}
\label{eq:rec2}
D_2' (K-\alpha)(\beta -1) \leq 2\alpha D_2 \left(\frac{\noise_t}{\rstep{t} + K-1}\right)^\alpha
\end{align}
Since $ \beta = 1 + 2 \alpha/(K-\alpha)$, the left hand side of \eqref{eq:rec2} equals $2\alpha D_2'$ and we conclude that $D_2' \leq D_2 (\noise_t / (\rstep{t} + K -1))^\alpha$.
Applying Theorem \ref{thm:aw_slow} we get:
\begin{equation*}
 h_t \leq  D_2'\left(\frac{\noise_t}{\rstep{t} + K-1}\right)^{\alpha}\leq D_2 \left(\frac{\noise_t}{\rstep{t} + K-1}\right)^{2\alpha}
\end{equation*}
The induction step is completed by observing that $\rstep{t} - 1 = \rstep{t-1}$. 
The initialization is easily verified for $t=2$.
If $h_t=0$, then by Lemma \ref{lem:ht_gtaw} yields $g_t^{\rm AW} = 0$
and the induction is treated as \textbf{Case 1}.


\subsection{Proof of Theorem \ref{thm:aw_slow}}
 \label{sec:pf_aw_slow}
We proceed by induction and assume for some $t>0$ that
$h_t  \leq D_2' (\noise_t/(\rstep{t-1}+K))^{\alpha}$ holds.
First of all, observe that from the $L$-smoothness of $f(\prm)$,
\begin{equation}
h_{t+1} \leq h_t + \hat{\gamma}_t \langle \nabla f(\prm_t) , {\bm d}_t \rangle
+ \frac{1}{2} \hat{\gamma}_t^2 \Curve \eqs.
\end{equation}
Moreover, we have:
\begin{align}
\langle \nabla f(\prm_t) , {\bm d}_t \rangle
& =  \langle \hat{\nabla} \Obj (\prm_t)
 , {\bm d}_t \rangle - \langle \bm{\epsilon}_t , {\bm d}_t \rangle \leq \langle \hat{\nabla} \Obj (\prm_t) , \atom_t^{\sf FW} - \prm_t \rangle
- \langle \bm{\epsilon}_t , {\bm d}_t \rangle  \nonumber \\
& \leq \langle \hat{\nabla} \Obj (\prm_t) , \prm_\star - \prm_t \rangle
- \langle \bm{\epsilon}_t , {\bm d}_t \rangle = \langle \grd \Obj (\prm_t) , \prm_\star - \prm_t \rangle
+ \langle \bm{\epsilon}_t , \prm_\star - \prm_t - {\bm d}_t \rangle \nonumber \\
& \leq - h_t + 2 \rho \| \bm{\epsilon}_t \|
\end{align}
where we used the condition of line \ref{ln:cond} (Algorithm~\ref{alg:ofw_away}) in the first inequality and the fact $\| \prm_\star - \prm_t - {\bm d}_t \|_\star \leq 2 \rho$ in the last inequality.
This gives
\begin{equation} \label{eq:ind_slow}
h_{t+1} \leq (1 - \hat{\gamma}_t)h_t
+ 2 \hat{\gamma}_t\rho \sigma \Big(\frac{\noise_t}{K+\rstep{t-1}} \Big)^\alpha + \frac{1}{2} \hat{\gamma}_t^2 \Curve \eqs,
\end{equation}
where we have used H\ref{iii1} and the fact that $\rstep{t-1} \leq t - 1$.

Consider the two cases: if a drop step (line~\ref{ln:drop}) is taken at iteration $t+1$,
the following result that is analogous to Lemma \ref{lem:drop} gives the induction.
\begin{Lemma}
\label{lem:drop_slow}
Suppose that $h_t \leq D_2' (\noise_t / (K+\rstep{t-1}))^{2\alpha}$
for $\alpha \in (0,1]$,
and that a drop step is taken at time $t+1$ (see Algorithm \ref{alg:ofw_away} line \ref{ln:drop}), then
\begin{equation}
 h_{t+1} \leq D_2'  \left( \frac{\noise_{t+1}}{K+\rstep{t}} \right)^{\alpha} \eqs.
\end{equation}
\end{Lemma}
\begin{proof}
 See \autoref{sec:pf_drop_slow}.
\end{proof}

On the other hand, if a drop step is \emph{not} taken,
notice that we will have $\hat{\gamma}_t = \gamma_{\rstep{t}} = K/(K+\rstep{t}-1)$ and $\rstep{t} = \rstep{t-1} + 1$.
Consequently, the same induction argument in
\autoref{sec:pf_converge} (replacing $t$ by $\rstep{t}$ and
consider $h_{t+1} - D_2' ( \noise_{t+1} / (K+\rstep{t}))^\alpha$) shows:
\begin{equation}
 h_{t+1} \leq D_2'  \left( \frac{\noise_{t+1}}{K+\rstep{t}} \right)^{\alpha}.
\end{equation}
The initialization of the induction is easily checked for $t=2$.


\subsection{Proof of Lemma \ref{lem:drop}}
\label{sec:pf_drop}
Since iteration $t+1$ is a drop step, we have by construction (\autoref{alg:ofw_away} line \ref{ln:drop})
\begin{equation*}
 \hat{\gamma}_t = \gamma_{\max} \leq \frac{K}{K + \rstep{t}} \quad \text{and} \quad \rstep{t} = \rstep{t-1} \eqs.
\end{equation*}
From \eqref{eq:dyn} and the assumption in the lemma, we consider two cases:
if $\sqrt{h_t} - \hat{\gamma}_t \sqrt{\mu \daw^2 / 2} \leq 0$, then we have
\begin{equation} \begin{split}
h_{t+1} - D_2 \left( \frac{\noise_{t+1}}{K+\rstep{t}}\right)^{2\alpha} &
\leq 2 \hat{\gamma}_t \rho \sigma \Big( \frac{\noise_t}{\rstep{t-1}+K} \Big)^\alpha + \frac{1}{2}
\Curve \hat{\gamma}_t^2  - D_2 \left( \frac{\noise_{t+1}}{\rstep{t}+K}\right)^{2\alpha} \\
& \hspace{-1.5cm} \leq 2 \rho \sigma \frac{K \cdot (\noise_{t+1})^\alpha }{(\rstep{t}+K)^{1+\alpha}}
+ \frac{1}{2} \Curve \left( \frac{K}{\rstep{t}+K}  \right)^2 - D_2 \left( \frac{\noise_{t+1}}{\rstep{t}+K}\right)^{2\alpha} \\
& \hspace{-1.5cm} \leq \left( \frac{\noise_{t+1}}{\rstep{t}+K}\right)^{2\alpha} \Big( 2\rho \sigma K + K^2 \Curveaw / 2 - D_2 \Big)
\end{split} \end{equation}
The second inequality is due to $\rstep{t} = \rstep{t-1}$ and $\hat{\gamma}_t = \gamma_{max} \leq K / (K+\rstep{t})$. The last inequality is due to $2 \alpha \leq \min\{2, 1 + \alpha\}$ for all $\alpha \in (0,1]$ and $\noise_t$ is an increasing sequence with $\noise_t \geq 1$. It can be verified that the right hand side is non-positive using the definition of $D_2$.

On the other hand, if $\sqrt{h_t} - \hat{\gamma}_t \sqrt{\mu \daw^2 / 2} > 0$, we have from \eqref{eq:dyn}
\begin{align*}  
&h_{t+1} - D_2 \left( \frac{\noise_{t+1}}{\rstep{t}+K}\right)^{2\alpha} \\ 
&\leq
\sqrt{h_t} \Big( \sqrt{h_t} - \hat{\gamma}_t \sqrt{\mu \daw^2 / 2} \Big) +
\frac{1}{2} \Curveaw \hat{\gamma}_t^2 +
2 \hat{\gamma}_t \rho \sigma \Big( \frac{\noise_t}{\rstep{t-1}+K} \Big)^\alpha
- D_2 \left( \frac{\noise_{t+1}}{\rstep{t}+K}\right)^{2\alpha} \\
&  \leq \frac{1}{2} \Curveaw \hat{\gamma}_t^2 +
2 \hat{\gamma}_t \rho \sigma \Big( \frac{\noise_t}{\rstep{t-1}+K} \Big)^\alpha
- \hat{\gamma}_t \sqrt{D_2 \mu \daw^2 / 2}  \left( \frac{\noise_t}{\rstep{t-1}+K}\right)^{\alpha} \\
&  = \hat{\gamma}_t \Big( \frac{1}{2} \Curveaw \hat{\gamma}_t
+ 2 \rho \sigma \Big( \frac{\noise_t}{\rstep{t}+K} \Big)^\alpha
- \sqrt{D_2 \mu \daw^2 / 2}  \left( \frac{\noise_t}{\rstep{t}+K}\right)^{\alpha}  \Big) \\
& \leq \hat{\gamma}_t \Big( \frac{K \Curveaw / 2 }{\rstep{t} + K}
+ \Big( 2 \rho \sigma
- \sqrt{D_2 \mu \daw^2 / 2} \Big) \left( \frac{\noise_t}{\rstep{t}+K}\right)^{\alpha}  \Big) \\
&  \leq \hat{\gamma}_t \Big( \frac{\noise_t}{\rstep{t}+K} \Big)^\alpha \Big( K \Curveaw / 2 + 2 \rho \sigma -  \sqrt{D_2 \mu \daw^2 / 2}  \Big) \eqs.
\end{align*}
The last inequality is due to $\alpha \leq 1$. Similarly, by the definition of $D_2$, we observe that the RHS in the above inequality is non-positive.


\subsection{Proof of Lemma \ref{lem:drop_slow}}
\label{sec:pf_drop_slow}
Using \eqref{eq:ind_slow} gives the following chain
\begin{equation}
\begin{split}
 h_{t+1} - D_2'  \left( \frac{\noise_{t+1}}{K+\rstep{t}} \right)^{\alpha}
& \leq (1-\hat{\gamma}_t) h_t + 2 \hat{\gamma}_t \rho \sigma \left( \frac{\noise_{t+1}}{K+\rstep{t}} \right)^\alpha + \frac{1}{2} \Curve \hat{\gamma}_t^2 - D_2'  \left( \frac{\noise_{t+1}}{K+\rstep{t}} \right)^{\alpha} \\
& \hspace{-2cm} \leq (1-\hat{\gamma}_t) D_2'  \left( \frac{\noise_{t}}{K+\rstep{t}} \right)^{\alpha} + 2 \hat{\gamma}_t \rho \sigma \left( \frac{\noise_{t+1}}{K+\rstep{t}} \right)^\alpha + \frac{1}{2} \Curve \hat{\gamma}_t^2 - D_2'  \left( \frac{\noise_{t+1}}{K+\rstep{t}} \right)^{\alpha} \\
& \hspace{-2cm} \leq \hat{\gamma}_t \left( (-D_2' + 2 \rho \sigma)  \left( \frac{\noise_{t}}{K+\rstep{t}} \right)^{\alpha} + \hat{\gamma}_t \frac{\Curve}{2} \right) \\
& \hspace{-2cm} \leq \hat{\gamma}_t \left( \Big(-D_2' + 2 \rho \sigma + \frac{1}{2} K \Curve \Big)  \left( \frac{\noise_{t}}{K+\rstep{t}} \right)^{\alpha} \right) \leq 0 \eqs.
\end{split}
\end{equation}
In the above, the second inequality is due to $1-\hat{\gamma}_t \geq 0$
and the induction hypothesis; the third inequality is due to $\noise_t$ is increasing
and; the last inequality is due to $\hat{\gamma}_t < K /(K + \rstep{t})$.
The proof is completed.


\section{Fast convergence of O-AW without strong convexity}
\label{sec:pf_nonstrong}

The proof is based on a generalization of Lemma \ref{lem:ht_gtaw}, and the
following result is borrowed from Theorem 11 in
\cite{Lacoste_Jaggi15}.

We focus on the anytime/regret bound studied in Section 3.1 below.
In particular, the relaxed conditions for a regret bound of ${\cal O}(\log^3 T /T)$ and anytime bound
of ${\cal O} (\log^2 t / t)$ are that \emph{(i)} $\Cset$ is a polytope and \emph{(ii)} the
loss function can be written as:
\begin{equation}
f(\prm) = g( {\bm A} \prm ) + \langle {\bm b}, \prm \rangle \eqs.
\end{equation}
where $g$ is $\mu_g$-strongly convex.
For a general matrix ${\bm A}$,  $f(\prm)$ may not be strongly convex.

Define ${\bm C}$ to be the matrix with rows containing
the linear inequalities defining $\Cset$.
Let $c_h$ be the Hoffman constant \cite{Lacoste_Jaggi15} for the matrix
$[{\bm A}; {\bm b}^\top; {\bm C}]$, $G = \max_{\prm \in \Cset} \| \grd g( {\bm A} \prm ) \|$
be the maximal norm of gradient of $g$ over ${\bm A} \Cset$, $\rho_{\bm A}$
be the diameter of ${\bm A} \Cset$ and we define the generalized strong convexity constant:
\begin{equation}
\tilde{\mu} \eqdef \frac{1}{ 2 c_h^2 ( \| {\bm b} \| M + 3G \rho_{\bm A} + (2/\mu_g) (G^2+1) } \eqs.
\end{equation}
Under H\ref{aw1} and assuming that $h_t > 0$ holds, applying the inequality (43) from \cite{Lacoste_Jaggi15} yields
\begin{equation} \label{eq:gtns}
\bar{g}_t^{\sf AW} \geq \delta_{AW} \sqrt{ 2 \tilde{\mu} \cdot h_t } \eqs.
\end{equation}
Subsequently, the ${\cal O}(\log^2 T / T)$ anytime bound and ${\cal O}(\log^3 T /T)$ regret bound in
Theorem \ref{cor:anytime} can be obtained by
repeating the proof in \autoref{sec:converge2} with \eqref{eq:gtns}.

\section{Improved gradient error bound for online MC} \label{prop:mc_simp}
Our goal is to show that with high probability,
\beq
\| \grd F_t(\prm) - \grd f(\prm) \|_{\sigma,\infty} = {\cal O} (\sqrt{\log t / t }),~\forall~ t~\text{sufficiently large}.
\eeq

To facilitate our proof, let us state the following conditions on the observation noise statistics:
\begin{assumptionb}
\label{A1}
The noise variance is finite, that is
there exists a constant
$\sigup >0$ such that for all $\vartheta \in \RR$,
$
0 \leq \lgp''(\vartheta)  \leq \sigup^2  \eqsp,
$
and the noise is sub-exponential \ie
there exist a constant $\sexp \geq 1$ such
that for all $(k,l) \in [m_1] \times [m_2]$:
\begin{equation}
\label{eq:def_sub}
\int\exp\left(\lambda^{-1}\left|y - \lgp'(\tX_{k,l})\right|\right)p_{\tX}(y|k,l)dy \leq \rme \eqs,
\end{equation}
where $p_{\tX}(\cdot)$
is defined as $p_{\tX}( y | k, l ) \eqdef
 \bms(y) \exp\left(y \tX_{k,l} - \lgp(\tX_{k,l}) \right)$ and
$\rme$ is the natural number.
\end{assumptionb}
\begin{assumptionb}
\label{A2}
There exists a finite constant $\mrc>0$
such that for all $\prm \in \contset$, $k \in [m_1]$, $l \in [m_2]$
\begin{equation}
 \mrc \geq \max
 \left( \sqrt{\sum_{l=1}^{m_2} \lgp'(\prm_{k,l})^2},
 \sqrt{\sum_{k=1}^{m_1} \lgp'(\prm_{k,l})^2} \right) \eqs.
\end{equation}

Notice that $\mrc = {\cal O}( \sqrt{ \max\{ m_1, m_2 \} } )$.

\end{assumptionb}
We remark that A\ref{A1} and A\ref{A2} are satisfied by all the
exponential family distributions.
We also need the following proposition.

\begin{Prop}
\label{prop:bernstein_exp}
 Consider a finite sequence of independent
 random matrices $(\Bern{Z_{s}})_{1 \leq s \leq t}\in \RR^{m_1 \times m_2}$
 satisfying $\EE[\Bern{Z_{i}}]=0$. For some $U>0$,
 assume
 \begin{equation}
   \inf \{ \sexp >0: \EE [ \exp (\| \Bern{Z_{i}} \|_{\sigma,\infty}/ \sexp) ] \leq \rme \}
   \leq U \quad \forall i \in [n],
 \end{equation}
 and there exists $\sigma_Z$ s.t.
 \begin{equation}
 \sigma^2_Z \geq \max \left\{\left\| \frac{1}{t} \sum_{s=1}^{t}
 \EE[  \Bern{{Z_s}}\Bern{Z_{s}^\top} ] \right\|_{\sigma,\infty},
 \left\| \frac{1}{t}\sum_{s=1}^{t} \EE[ \Bern{Z_{s}}^\top\Bern{Z_{s}} ]\right\|_{\sigma,\infty}\right\} \eqs.
\end{equation}
Then for any $\nu>0$,
 with probability at least $1-\rme^{-\nu}$
 \begin{equation}
  \norm{\frac{1}{t} \sum_{i=1}^{t} \Bern{Z_{i}}}_{\sigma,\infty}
  \leq c_U \max \left\{ \sigma_Z \sqrt{ \frac{\nu + \log(d)}{t}} ,
  U \log(\frac{U}{\sigma_Z})\frac{\nu + \log(d)}{t} \right \} \eqs,
 \end{equation}
with $c_U$ an increasing constant with $U$.
\end{Prop}

\begin{proof}
This result is proved in Theorem 4 in \cite{koltchinskii13}
for symmetric matrices. Here we state a slightly different result
because $\sigma^2_Z$ is an upper bound of the variance and
not the variance itself. However, it does not the alter the proof
and the result stays valid. This concentration is extended to rectangular matrices by dilation,
see Proposition~11 in \cite{Klopp14}
for details.
\end{proof}

Our result is stated as follows.

\begin{Prop}
\label{prop:gradient_conc}
Assume A\ref{A1}, A\ref{A2}
and that the sampling distribution is uniform.
Define
the approximation error
$\bm{\epsilon}_t (\prm) \eqdef \grd F_t( \prm ) - \nabla f(\prm)$.
With probability at least $1-  \epsilon$,
for any
$t \geq T_\epsilon \eqdef  (\sexp/\sigup)^2 \log^2(\sexp/\sigup)\log(d+2d/\epsilon)$,
and any $\prm \in \contset_\trad$:
\begin{equation*}
\label{eq:grad_dev}
 \normop{\bm{\epsilon}_t (\prm)} = \mathcal{O} \left( c_{\sexp } \left( \mrc
  + \sigup  \right)
  \sqrt{ \frac{\log(d(1 + t^2 /\epsilon)) }{t(m_1 \wedge  m_2)}} \right) \eqs,
\end{equation*}
with
$\|\cdot\|_{\sigma, \infty}$ the
operator norm, $c_\sexp$ a constant which depends only on $\sexp$. The constants $\sexp$, $\sigup$ and $\mrc$ are
defined in A\ref{A1} and A\ref{A2}.
\end{Prop}

\begin{proof}
For a fixed $\prm$, by the triangle inequality
\begin{multline*}
  \normop{ \bm{\epsilon}_t (\prm) }  \leq
  \normop{\frac{1}{t} \sum_{s=1}^t \Y_se_{k_s}e^{'\top}_{l_s} - \EE[\Y_se_{k_s}e^{'\top}_{l_s}] }  + \normop{ \frac{1}{t} \sum_{s=1}^t \lgp'(\prm_{k_s,l_s})e_{k_s} e^{'\top}_{l_s}  -
 \EE[ \lgp'(\prm_{k_s,l_s})e_{k_s} e^{'\top}_{l_s}]}
\end{multline*}
Define
$Z_s \eqdef \Y_s e_{k_s}e^{'\top}_{l_s} - \EE[\Y_s e_{k_s}e^{'\top}_{l_s}]$, then
\begin{align*}
 &\normop{\EE[Z_s Z_s^\top]}  \leq \normop{\EE[\Y_s^2 e_{k_s}e^{'\top}_{l_s}e^{'}_{l_s}e_{k_s}^\top]} \eqs, \\
 &= \normop{\frac{1}{m_1m_2} \diag \left( \left(\sum_{l=1}^{m_2} \EE[\Y_s^2|k,l] \right)_{k=1}^{m_1}\right)} \eqs,\\
 &= \frac{1}{m_1m_2}\max_{k \in [m_1]}
 \left(\sum_{l=1}^{m_2} \lgp''(\tX_{k,l}) + (\lgp'(\tX_{k,l}))^2 \right) \eqs,\\
%
&\leq \frac{\sigup^2}{m_1 \wedge m_2} + \frac{\mrc^2}{m_1  m_2} \leq \frac{\sigup^2 + \mrc^2}{m_1 \wedge m_2}\eqs,
\end{align*}
where we used the fact that the distribution belongs
to the exponential family for the second equality.
Similarly one shows that
$\normop{\EE[Z_s^\top Z_s]}$ satisfies the same upper bound.
Hence by Proposition \ref{prop:bernstein_exp} and A\ref{A1}, with probability at least $1 - \rme^{-\nu}$,
it holds
\begin{equation}
\normop{\frac{1}{t} \sum_{s=1}^t Z_s }  \leq
c_\sexp  \sqrt{\frac{(\sigup^2 + \mrc^2)(\nu + \log(d))}{t (m_1 \wedge  m_2)}}  \eqs,
\end{equation}
for $t$ larger than the threshold given in the proposition statement.
For the second term, define
$P_t \eqdef 1/t\sum_{s=1}^t e_{k_s}e_{l_s}^{'\top} - (m_1m_2)^{-1} {\bm 1  \bm 1^\top}$,
 we get
\begin{equation}
\begin{split}
\normop{ \frac{1}{t} \sum_{s=1}^t \lgp'(\prm_{k_s,l_s})e_{k_s}e^{'\top}_{l_s}
- \EE[\lgp'(\prm_{k_s,l_s})e_{k_s}e^{'\top}_{l_s}]}
=\normop{ P_t \odot (\lgp'( \prm_{k,l}))_{k,l}}
 \leq \mrc \normop{ P_t }  \eqs,
\end{split}
\end{equation}
where $\odot$ denotes the Hardamard product and we have used Theorem 5.5.3 in \cite{Horn_Johnson94} for the last inequality.
Define $Z'_s \eqdef e_{k_s}e_{l_s}^{'\top} - (m_1m_2)^{-1} {\bm 1  \bm 1^\top}$.
Since by definition, $\sexp \geq 1$, one can again apply Proposition \ref{prop:bernstein_exp}
for $U = \lambda$ and get with probability at least $1 - \rme^{-\nu}$,
\begin{equation}
\normop{ P_t} \leq
 c_{\sexp} \sqrt{ \frac{\nu + \log(d)}{t(m_1 \wedge  m_2)}} \eqs.
\end{equation}
Hence, by a union bound argument we find that
with probability at least $1 - 2\rme^{-\nu}$
\begin{equation}
  \normop{  \bm{\epsilon}_t }  \leq
  c_{\sexp} \left( 2\mrc
  + \sigup  \right)
  \sqrt{ \frac{\nu + \log(d)}{t(m_1 \wedge  m_2)}} \eqs.
\end{equation}
Taking $\nu = \log(1 + 2 t^2 /\epsilon)$ and applying a union bound argument yields the result.
\end{proof}

\section{Additional results: Online LASSO}\label{sec:olasso} \vspace{-.1cm}
Consider the setting where we are sequentially 
given i.i.d.~observations $({\bm Y}_t, {\bm A}_t)$ such that ${\bm Y}_t \in \RR^m$ is
the response, ${\bm A}_t \in \RR^{m \times n}$ is the random design
and
\begin{equation} \label{eq:sparse_model}
{\bm Y}_t = {\bm A}_t \bar{\prm} + {\bm w}_t\eqs,
\end{equation}
where the vector ${\bm w}_t$ is i.i.d., $[{\bm w}_t]_i$ is independent of $[{\bm w}_t]_j$ for $i \neq j$ 
and $[{\bm w}_t]_i$ is zero-mean and sub-Gaussian with parameter $\sigma_w$. 
We suppose that the unknown parameter $\bar{\prm}$ is sparse. 
Attempting to learn $\bar{\prm}$, a natural choice for 
the loss function at round $t$ is the square loss, i.e.,
\beq
f_t( \prm) = (1/2) \| {\bm Y}_t - {\bm A}_t \prm \|_2^2
\eeq
and the stochastic cost associated is 
$\Obj(\prm) \eqdef \frac{1}{2} \EE_{\bar{\prm}} [ \| {\bm Y}_t - {\bm A}_t \prm \|_2^2 ]$.
As $\bar{\prm}$ is sparse, the constraint set is designed to be the $\ell_1$ ball, i.e.,
$\Cset = \{ \prm \in \RR^n ~:~ \| \prm \|_1 \leq r \}$, where $r > 0$ is a regularization constant. 
Note that $\Cset$ is a polytope. 

The aggregated gradient can be expressed as
\beq
\grd F_t( \prm_t ) = t^{-1} \Big( \sum_{s=1}^t {\bm A}_s^\top {\bm A}_s \Big) \prm_t - t^{-1} \Big( \sum_{s=1}^t {\bm A}_s^\top {\bm Y}_s \Big) \eqs.
\eeq
Similar to the case of online matrix completion, the terms $\sum_{s=1}^t {\bm A}_s^\top {\bm A}_s$ 
and $\sum_{s=1}^t {\bm A}_s^\top {\bm Y}_s$ can be computed `on-the-fly' as running sums. 
Applying O-FW (\autoref{alg:ofw}) or O-AW (\autoref{alg:ofw_away}) with the
above aggregated gradient yields an online LASSO algorithm
with a constant complexity (dimension-dependent) per iteration. 
Notice that as $\Cset$ is an $\ell_1$ ball constraint, the linear optimization in Line~\ref{ofw:lo} 
of \autoref{alg:ofw} or \eqref{eq:away} in \autoref{alg:ofw_away} can be evaluated simply as 
$\atom_t = -r \cdot {\rm sign}([ \grd F_t (\prm_t) ]_i ) \cdot {\bm e}_i$, 
where $i = \arg \max_{j \in [n]} | [\grd F_t (\prm_t) ]_j |$.

Similar to the case of online MC, we derive the following ${\cal O} ( \sqrt{ \log t / t } )$ bound for the 
gradient error:
\begin{Prop} \label{prop:lasso}
Assume that $\| {\bm A}_t^\top {\bm A}_t -  \EE [{\bm A}^\top {\bm A}]  \|_{max} \leq B_1$ and $ \| {\bm A}_t \|_{max} \leq B_2$ almost surely, with $\| \cdot \|_{max}$ being the matrix max norm.
Define $c \eqdef \max_{ \prm \in \Cset } \| \prm - \bar{\prm} \|_1$.
With probability at least $1- (1+1/n)(\pi^2 \epsilon / 6)$, the following holds for all $\prm \in \Cset$ and all $t \geq 1$:
\begin{equation} \label{eq:lassobd}
\| \grd F_t(\prm) - \grd f(\prm) \|_\infty \leq (c B_1 + \sqrt{m B_2\sigma_w^2}) \sqrt{ \frac{2 (\log(2n^2 t^2) - \log \epsilon)}{t} } \eqs,
\end{equation}
where $\| \cdot \|_\infty$ is the infinity norm and the dual norm of $\| \cdot \|_1$.
\end{Prop}

We observe that H\ref{iii1} is satisfied with $\noise_t$ asymptotically 
equivalent to $4 {\log (t) }$ and $\alpha = 0.5$. 
Furthermore, the stochastic cost $f$ is $L$-Lipschitz if $L {\bf I} \succeq \EE [ {\bm A}^\top {\bm A} ]$; 
$\mu$-strongly convex if $\EE [ {\bm A}^\top {\bm A} ] \succeq \mu {\bf 1}$ for 
some $\mu > 0$; 
and H\ref{aw1} is satisfied as $\Cset$ is a polytope. 
The analysis from the previous section applies, i.e., O-FW/O-AW has a regret bound of ${\cal O} ( \log^2 T / T)$ and an anytime bound of ${\cal O}( \log t / t)$. 

\begin{proof}
Notice that the gradient vector is given by:
\begin{equation}
\grd \Obj (\prm) = \EE [ {\bm A}^\top ({\bm A} \prm - {\bm Y}) ] = \EE [ {\bm A}^\top {\bm A}] \prm - \EE [ {\bm A}^\top {\bm Y}].
\end{equation}
We can bound the gradient estimation error as:
\begin{equation} \label{eq:ls1}
\begin{array}{l}
\displaystyle \| \grd F_t(\prm) - \grd f(\prm) \|_\infty \leq \Big\| \frac{1}{t} \sum_{s=1}^t {\bm A}_s^\top {\bm w}_s \Big\|_\infty + \Big\| \frac{1}{t} \sum_{s=1}^t \big( {\bm A}_{s}^\top {\bm A}_{s} - \EE [{\bm A}^\top {\bm A}]  \big) (\prm - \bar{\prm}) \Big\|_\infty 
\end{array} 
\end{equation}
To bound the second term in \eqref{eq:ls1}, we define ${\bm Z}_s \eqdef {\bm A}_{s}^\top {\bm A}_{s} - \EE [{\bm A}^\top {\bm A}] $. Observe that
\begin{equation}
\Big\| \frac{1}{t} \sum_{s=1}^t {\bm Z}_s (\prm - \bar{\prm}) \Big\|_\infty = \max_{i \in [n]} \Big| \frac{1}{t} \sum_{s=1}^t {\bm z}_{s,i} (\prm - \bar{\prm}) \Big|,
\end{equation}
where $ {\bm z}_{s,i}$ denotes the $i$th row vector in ${\bm Z}_s$. Furthermore, by the Holder's inequality,
\begin{equation}
\big| \frac{1}{t} \sum_{s=1}^t {\bm z}_{s,i} (\prm - \bar{\prm}) \big| \leq \| \prm - \bar{\prm} \|_1 \big\| \frac{1}{t} \sum_{s=1}^t {\bm z}_{s,i} \big\|_\infty,
\end{equation}
Now that ${\bm z}_{s,i}$ is a zero-mean, independent random vector with elements bounded in $[-B_1,B_1]$, applying the union bound and the Hoefding's inequality gives:
\begin{equation} \label{eq:ls2}
\PP \big( \big\|\textstyle  (1/t) \sum_{s=1}^t {\bm z}_{s,i} \big\|_\infty \geq x,~\forall~i \big) \leq 2n^2 \rme^{-\frac{x^2 t}{2B_1^2}}.
\end{equation}
Setting $x = B_1 \sqrt{2 (\log(2n^2 t^2) - \log \epsilon) / t}$ gives $\epsilon / t^2$ on the right hand side. With probability at least $1 - \epsilon / t^2$, we have 
\begin{equation} \label{eq:lsbound}
\Big\| \frac{1}{t} \sum_{s=1}^t {\bm Z}_s \prm \Big\|_\infty \leq c B_1 \sqrt{2 (\log(2n^2 t^2) - \log \Delta) / t},
\end{equation}

To bound the first term in \eqref{eq:ls1}, we find that the $i$th element of the vector ${\bm A}_s^\top {\bm w}_s$ is zero-mean. Furthermore, it can be verified that
\begin{equation}
\EE \big[ \rme^{\big( \lambda \sum_{j=1}^m A_{s,i,j} w_{s,j}\big)}\big] \leq \rme^{\lambda^2 \cdot m \sigma_w^2 B_2 / 2},
\end{equation}
for all $\lambda \in \RR$, where $A_{s,i,j} $ is the $(i,j)$th element of ${\bm A}_s$ and $w_{s,j}$ is the $j$th element of ${\bm w}_s$. In other words, the $i$th element of ${\bm A}_s^\top {\bm w}_s$ is sub-Gaussian with parameter $m \cdot \sigma_w^2 B_2$. It follows by the Hoefding's inequality that 
\begin{equation} \label{eq:ls3}
\PP \big( \big\| \frac{1}{t} \sum_{s=1}^t {\bm A}_s^\top {\bm w}_s \big\|_\infty \geq x \big) \leq 2n \rme^{-\frac{x^2 t }{2 m  B_2 \sigma_w^2}}.
\end{equation}
Setting $x = \sigma_w \sqrt{2 m B_2 (\log(2n^2 t^2) - \log \epsilon) / t}$ yields $\epsilon / (nt^2)$ on the right hand side. Combining \eqref{eq:ls2}, \eqref{eq:ls3} and using a union bound argument (for all $t \geq 1$) yields the desired result. 
\end{proof}

\subsection{Numerical Result}

We present numerical results on both synthetic data and realistic data. 

\textbf{Synthetic Data.} We set ${\bm A}_t = {\bm A}$ 
as fixed for all $t$ with dimension $80 \times 300$ and 
the parameter $\bar{\prm} \in \RR^{300}$ is a vector 
with $10 \%$ sparsity and independent ${\cal N}(0,1)$ elements. 
We also set $\sigma_w = 10$. The matrix ${\bm A}$ is generated 
as a random Gaussian matrix with 
independent ${\cal N}(0,1)$ elements.
For benchmarking purpose, we have compared the O-FW/O-AW's performance 
with a stochastic projected gradient (sPG) method \cite{stochastic_pg_lorenzo} 
with a fixed step size $1/L$. 

\begin{figure}[t]
\centering
 \includegraphics[width =0.35\textwidth]{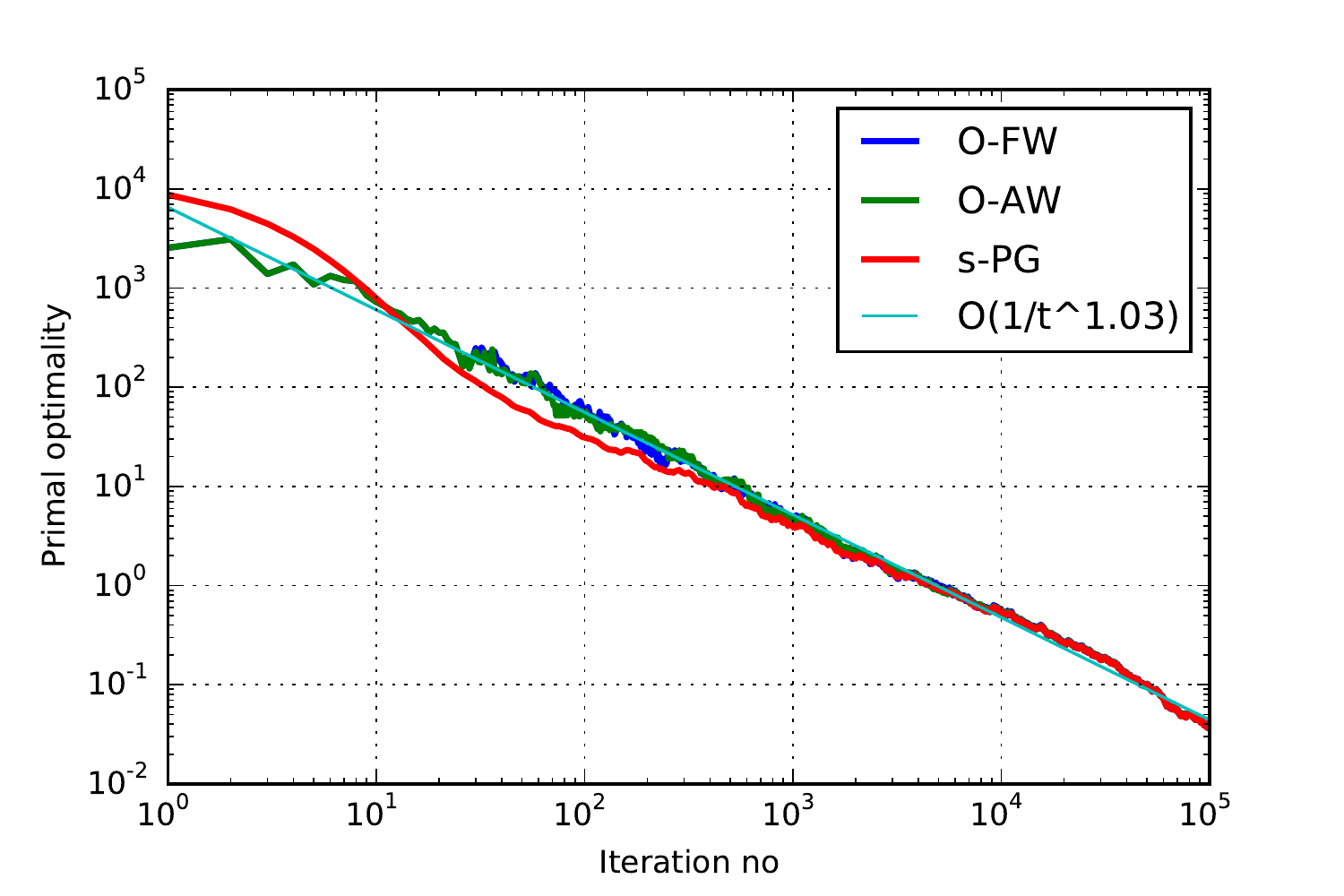}~
  \includegraphics[width =0.35\textwidth]{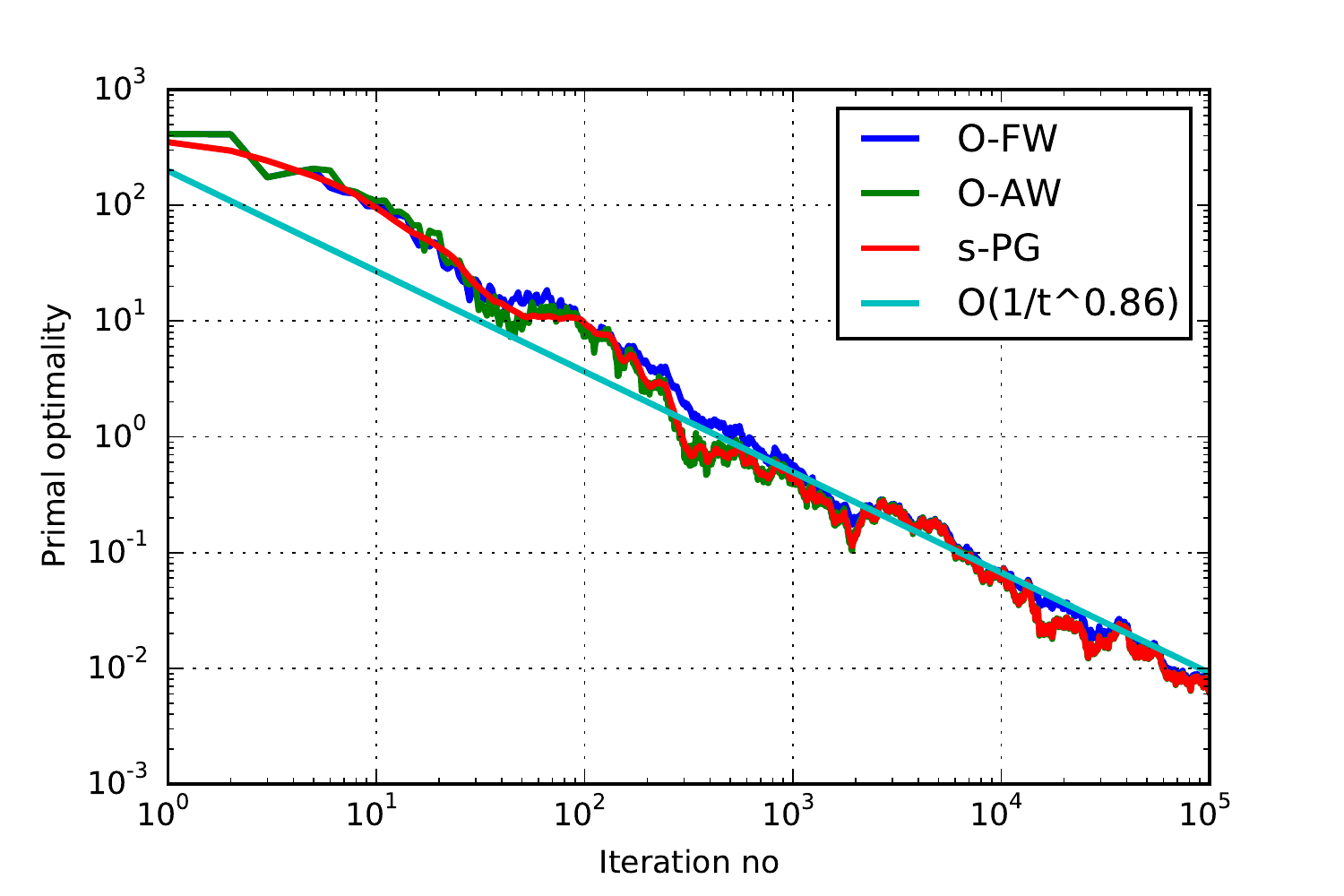}
\caption{\small Online LASSO with synthetic data. Convergence of the primal optimality for online LASSO with (Left) $r = 1.1 \| \bar{\prm} \|_1 > \| \prm^\star \|_1$; (Right) $r = 0.15 \| \bar{\prm} \|_1 = \| \prm^\star \|_1$.} 
\label{fig:lasso}
\end{figure}

\begin{figure}[t]
\centering
 \includegraphics[width =0.34\textwidth]{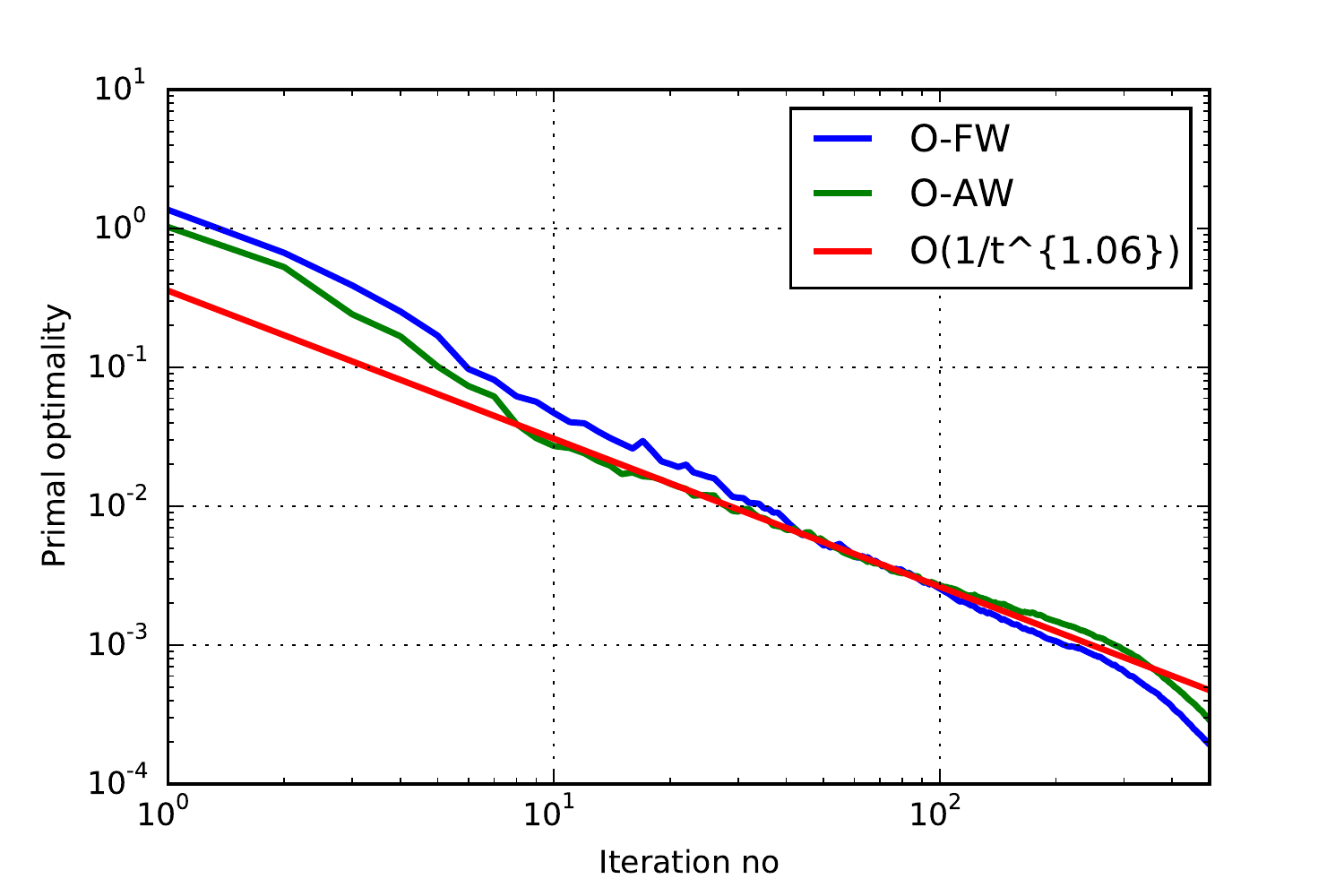} ~~~~
  \includegraphics[width =0.15\textwidth]{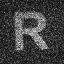}~~
\includegraphics[width =0.15\textwidth]{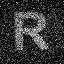} 
\caption{\small Online LASSO with single-pixel imaging data \texttt{R64.mat}. (Left) Convergence of the objective value. (Middle) Reconstructed image after $500$ iterations of O-FW; (Right) O-AW.}
\label{fig:largescale}
\end{figure}


\autoref{fig:lasso} plots the primal optimality
$h_t \eqdef \Obj (\prm_t) - \Obj (\prm^\star)$ with the round number $t$. 
The left figure corresponds to the scenario under H\ref{iii3i} as $\prm^\star$ belongs to the interior of $\Cset$. 
The simulation result corroborates with our analysis, 
which indicate a fast convergence rate of ${\cal O}(1/t)$. 
{In the right figure, we observe that although H\ref{iii3i} is not satisfied, 
the O-FW algorithm still maintains a 
convergence rate of $\sim {\cal O}(1/t)$, and O-AW is slightly outperforming O-FW. 
Examining the necessity of including H\ref{iii3i} in 
achieving a fast convergence rate for O-FW will be left for future investigation.} 
Lastly, the primal convergence rate of sPG is similar to O-FW. 
However, the per-iteration complexity of sPG is ${\cal O}(n \log n)$, 
while it is ${\cal O}(n)$ for the O-FW.

\textbf{Realistic Data.} We consider learning a sparse image $\prm$ from the dataset 
\texttt{R64.mat} available from \cite{onebitcs}. The dataset consists of $T = 4319$ 
one-bit measurements of a greyscale image of `R' with  size $64 \times 64$. 
The squared loss function is chosen such that $f_t (\prm) = (y_t - {\bm a}_t^\top \prm)^2$, where ${\bm a}_t \in \RR^n$ is a binary measurement vector and $n=4096$ is the vectorized image. 
For the O-FW/O-AW algorithms, 
we have (i) used batch processing by drawing a batch of $B = 5$ new observations and 
(ii) introduced an inner loop by repeating the O-FW/O-AW iterations, i.e., 
Line 4-5 of \autoref{alg:ofw} or Line 4-15 of \autoref{alg:ofw_away} for $50$ times 
within each iteration. 

As the optimal solution $\prm^\star$ is unavailable for this problem, \autoref{fig:largescale} compares the primal objective value $F_T(\prm_t)$ against the iteration number and the reconstructed image after $t_f = 500$ iterations of the tested algorithms. The figure shows that the convergence rates of these algorithms all converge at a rate of $\sim {\cal O}(1/t)$. 



{\small
\bibliographystyle{alpha}
\bibliography{references/references_all,references/references_all_2}
}

\end{document}